\documentclass[11pt,a4paper]{article}

\usepackage{subfig}

\usepackage{amssymb,amsmath,amsthm}
\usepackage[pdftex]{graphicx}
\graphicspath{{./figures/}}

\usepackage{float}

\usepackage[]{algorithm2e}

\makeatletter
\def\BState{\State\hskip-\ALG@thistlm}
\makeatother
\usepackage{authblk}

\usepackage[margin=0.8in]{geometry}

\usepackage[utf8]{inputenc} 
\usepackage[T1]{fontenc}    
\usepackage{hyperref}       
\usepackage{url,comment}            
\usepackage{booktabs}       
\usepackage{amsfonts}       
\usepackage{nicefrac}       

\newenvironment{myitemize}{
\begin{itemize}
 \setlength{\itemsep}{1pt}
 \setlength{\parskip}{0pt}
 \setlength{\parsep}{0pt}}{\end{itemize}
}

\title{Algorithmic Probability-guided Supervised Machine\\Learning on Non-differentiable Spaces}

\author[1,2]{Santiago Hern\'andez-Orozco}
\author[2,3,4,5]{Hector Zenil\thanks{Corresponding authors.}}
\author[2,4]{J\"urgen Riedel}
\author[4]{Adam Uccello}
\author[3,4]{Narsis A. Kiani}
\author[5]{Jesper Tegn\'er$^*$}
\affil[1]{Facultad de Ciencias, Universidad Nacional Aut\'onoma de M\'exico, Mexico City, Mexico.}
\affil[2]{ Oxford Immune Algorithmics, Oxford, U.K.}
\affil[3]{Algorithmic Dynamics Lab, Unit of Computational Medicine, Karolinska Institutet, Sweden. }
\affil[4]{Algorithmic Nature Group, LABORES, Paris, France.}
\affil[5]{King Abdullah University of, Science and Technology (KAUST), Computer, Electrical and Mathematical
Sciences and Engineering, Saudi Arabia.}

\begin{document}

\theoremstyle{plain}
\newtheorem{thm}{Theorem}
\newtheorem{lem}[thm]{Lemma}
\newtheorem{cor}[thm]{Corollary}
\newtheorem{prop}[thm]{Proposition}
\theoremstyle{definition}
\newtheorem{defn}[thm]{Definition}
\newtheorem{conj}[thm]{Conjecture}
\date{\vspace{-.5cm}}
\vspace{-.4cm}

\maketitle

\begin{abstract}

We show how complexity theory can be introduced in machine learning to help bring together apparently disparate areas of current research. We show that this new approach requires less training data and is more generalizable as it shows greater resilience to random attacks. We investigate the shape of the discrete algorithmic space when performing regression or classification using a loss function parametrized by algorithmic complexity, demonstrating that the property of differentiation is not necessary to achieve results similar to those obtained using differentiable programming approaches such as deep learning. In doing so we use examples which enable the two approaches to be compared (small, given the computational power required for estimations of algorithmic complexity). We find and report that (i) machine learning can successfully be performed on a non-smooth surface using algorithmic complexity; (ii) that parameter solutions can be found using an algorithmic-probability classifier, establishing a bridge between a fundamentally discrete theory of computability and a fundamentally continuous mathematical theory of optimization methods; (iii) a formulation of an algorithmically directed search technique in non-smooth manifolds can be defined and conducted; (iv) exploitation techniques and numerical methods for algorithmic search to navigate these discrete non-differentiable spaces can be performed; in application of the (a) identification of generative rules from data observations; (b) solutions to image classification problems more resilient against pixel attacks compared to neural networks; (c) identification of equation parameters from a small data-set in the presence of noise in continuous ODE system problem, (d) classification of Boolean NK networks by (1) network topology, (2) underlying Boolean function, and (3) number of incoming edges.\\



\end{abstract}

\section{Introduction}

Given a labelled data-set, a loss function is a mathematical construct that assigns a numerical value to the discrepancy between a predicted model-based outcome and its \textit{real outcome}. A cost function aggregates all losses incurred into a single numerical value that, in simple terms, evaluates how close the model is to the \textit{real data}. The goal of minimizing an \textit{appropriately formulated} cost function is ubiquitous and central to any machine learning algorithm. The main heuristic behind most training algorithms is that fitting a sufficiently representative training set will result in a model that will capture the \textit{structure} behind the elements of the target set, where a model is fitted to a set when the absolute minimum of the cost function is reached. 

The algorithmic loss function that we introduce is designed to quantify the discrepancy
between an inferred program (effectively a computable model of the data) and the data.

Algorithmic complexity~\cite{solo,kolmo, chaitin}, along with its associated complexity function $K$, is the accepted mathematical definition of randomness. Here, we adopt algorithmic randomness---with its connection to algorithmic probability---to formulate a universal search method~\cite{levin1974laws,solomonoff1986application} for exploring non-entropy-based loss/cost functions in application to AI, and to supervised learning in particular. We exploit novel numerical approximation methods based on algorithmic randomness to navigate undifferentiable problem representations capable of implementing and comparing local estimations of algorithmic complexity, as a generalization of particular entropy-based cases, such as those rooted in cross entropy or KL divergence, among others. 

In~\cite{zenilthesis,d5} and ~\cite{plos,bdm}, a family of numerical methods was introduced for computing lower bounds of algorithmic complexity using algorithmic probability. The algorithmic complexity $K_L$ of an object $x$ is the length of the shortest binary computer program $p$, that running on a Turing-complete language $L$, can reproduce $x$ and halt. That is, $K_L(x)= min_p\{|p|:L(p)=x\}.$ The Invariance theorem~\cite{solo,kolmo,chaitin} guarantees that the choice of computer language $L$ has only an impact bounded by a constant that can be thought of as the length of the compiler program needed to translate a computer program from one computer language into another.  
The algorithmic probability~\cite{solo,levin} of an object $x$ is the probability $AP$ of a binary computer program $p$ producing $x$ by chance (i.e. considering that keystrokes are binary instructions) running on a Turing-complete computer language $L$ and halting. That is, $$AP(x) := \sum_{p:L(p)=x} \frac{1}{|p|}\sim K(x).$$ Solomonoff and Levin show that $AP$ is an optimal computable inference method~\cite{solo2} and that any other inference method is either a special case less powerful than $AP$, or indeed is $AP$ itself~\cite{levin}. Algorithmic probability is related to algorithmic complexity by the so-called Coding theorem: $K(x) \sim -\log_2 AP(s)$. 

The Coding theorem~\cite{d5,plos} and Block Decomposition methods~\cite{bdm} provide a procedure to navigate the space of computable models matching a piece of data, allowing the identification of sets of programs sufficient to reproduce the data regardless of its length, and thus relaxing the minimal length requirement. In conjunction with classical information theory, these techniques constitute a hybrid, divide-and-conquer approach to universal pattern matching, combining the best of both worlds in a hybrid measure (BDM). The divide-and-conquer approach entails the use of an unbiased library of computable models, each capturing small segments of the data. These explore and build an unbiased library of computable models that can explain small segments of a larger piece of data, the conjoined sequence of which can reproduce the whole and constitute a computable model--as a generalization of statistical approaches typically used 
in current approaches to machine learning.

Interestingly, the use of algorithmic probability and information theory to define AI algorithms has theoretically been proposed before \cite{solomonoff1986application,solomonoff2003kolmogorov}. Yet, it has received limited attention in practice compared to other less powerful but more accessible techniques, due to the theoretical barrier that has prevented researchers from exploring the subject further. However, in one of his latest interviews, if not the last one, Marvin Minsky suggested that the most important direction for AI was actually the study and introduction of Algorithmic Probability~\cite{minsky}.

\section{An Algorithmic Probability Loss Function}\label{lossS}

The main task of a loss function is to measure the discrepancy between a value predicted by the model and the actual value as specified by the training data set. In most currently used machine learning paradigms this discrepancy is measured in terms of the differences between numerical values, and in case of cross-entropy loss, between predicted probabilities.  Algorithmic information theory offers us another option for measuring this discrepancy--in terms of the \textit{algorithmic distance or information deficit} between the predicted output of the model and the real value, which can be expressed by the following definition:


\begin{defn} \label{aLoss}
Let $y$ be the real value and $\hat{y}$ the predicted value. The \emph{algorithmic loss function}
$L_a$ is defined as $L_a(y,\hat{y}) = K(y|\hat{y}).$ 
It can be interpreted as \textit{the loss incurred by the model at data sample $(x_i,y_i)$, and is defined as the information deficit between the real value with respect to the predicted value.}
\end{defn}
There is a strong theoretical argument to justify the Def.~\ref{aLoss}. Let's recall that given a training set $X = \langle x_i, y_i \rangle$, an AI model $M$ aims to capture, as well as possible, the underlying rules or mechanics that associate each input $x_i$ with its output $y_i$. Let's denote by $M^*$ the \textit{perfect} or \textit{real} model, such that $M^*(x_i)=y_i$.  It follows that an \textit{ideal} optimization metric would measure \textit{how far our model $M$ is from $M^*$}, which in algorithmic terms is denoted by $K(M^*|M)$. However, we do not have access to much information regarding $M^*$ itself. What we do have is a set of pairs of the form $(x_i, y_i=M^*(x_i))$. Thus the problem translates into minimizing the distance between a program $M^*$ that outputs $y_i$ and a program $M$ that outputs $\hat{y}_i$ given $x_i$. Now, given that $x_i$ is constant for $M$ and $M^*$, the objective of an optimization strategy can be interpreted as minimizing  $K(y_i|\hat{y_i})$ for all $y_i$ in our data sets. Therefore, with the proposed algorithmic loss function, we are not only measuring how far our predictions are from the real values, but also how \textit{far our model} is from the real explanation behind the data in a fundamental algorithmic sense.

An algorithmic \emph{cost function} must be defined as a function that aggregates the algorithmic loss incurred over a supervised data sample. At this moment, we do not have any reason, theoretical or otherwise, to propose any particular loss aggregation strategy. As we will show in subsequent sections, considerations such as continuity, smoothness and differentiability of the cost function are not applicable to the algorithmic cost function. We conjecture that any aggregation technique that \textit{correctly and uniformly} weights the loss incurred through all the samples will be equivalent, the only relevant considerations being training efficiency and the statistical properties of the data.  However, in order to remain congruent with the most widely used cost functions, we will, for the purpose of illustration, use the \emph{sum of the squared algorithmic differences} $$J_a(\hat{X},M) =\sum_ {(x_i,y_i) \in \hat{X}} K(y_i|M(x_i))^2.$$

\section{Categorical Algorithmic Probability Classification}

One of the main fields of application for automated learning is the categorical classification of objects. These classification tasks are often divided into \textit{supervised} and \textit{unsupervised} problems. In its most basic form, a supervised categorical classification task can be defined, given a set of objects $X=\{x_1,\dots,x_i,\dots,x_n\}$ and a set of finite categories $C=\{c_1,\dots,c_j,\dots,c_m\}$, as that of finding a computable function or model $M:X \rightarrow C$ such that $M(x_j)=c_j$ if and only if $x_i$ \textit{belongs to} $c_j$ according to  \textit{previously agreed criteria}. In this section we  apply our \textit{hybrid machine learning} approach to supervised classification tasks.

Now, it is important to note that it is not constructive to apply the algorithmic loss function (Def.~\ref{aLoss}) to the abstract representations of classes that are commonly used in machine learning classifiers. For instance, the output of a softmax function is a vector of probabilities that represent how likely it is for an object to belong to each of the classes (\cite{bishop1995neural}), which is then assigned to a one-hot vector that represents the class itself. However, the integer that such a one-hot vector signifies is an abstract representation of the class that was arbitrarily assigned, and therefore has little to no algorithmic information pertaining to the class itself. 
Accordingly, in order to apply the algorithmic loss function to classification tasks, we need to seek a model that outputs an \textit{information rich object} that can be used to identify the class regardless of the context within which the problem was defined. In other words, the model must output the \textit{class itself} or a similar enough object that identifies it.

We can find the needed output in the definition of the classification problem: a class $c_j$ is defined as all the $x_{i_j}$ that are associated with it, and finding any underlying regularities to define a class beyond that is the task of a machine learning model. It follows that an algorithmic information model must output an object that \emph{minimizes the algorithmic distance to all the members of the class}, so that we can classify them. Therefore, a correct interpretation of the \textit{general definition} of the algorithmic loss function for a model $M$ is the following:
\begin{equation}\label{aCE}
L_a(y_i,\hat{y_i}) = K(x_i|M(x_i)),
\end{equation}
\noindent{}where  $y_i$ is the class to which $x_i$ belongs, while $M(x_i)=\hat{y_i}$ is the output of the model. What the equation \ref{aCE} is saying is that the model must produce as an output an object that is algorithmically close to all the elements of the class. In unsupervised classification tasks this object is known as a \textit{centroid of a cluster}~\cite{lloyds}. This means that the algorithmic loss function is inducing us to universally define algorithmic probability classification as a clustering by an algorithmic distance model.\footnote{An unlabelled classificatory algorithmic information schema has been proposed by \cite{cilibrasi2005clustering}.} 
A general schema in our algorithmic classification proposal is based on the concept of a nearest centroid classifier. 
\begin{defn}\label{clas}
Given a training set $\hat{X}=\{(x_1,c_{j_1}),\dots,(x_i,c_{j_i}),\dots,(x_n,c_{j_n})\}$ with $m$ different $c_j$ classes, an algorithmic classification model consists of a set of centroids $C^*=\{c^*_1,\dots,c^*_j\dots,c^*_m\}$ such that each minimizes the equation $$J_a(\hat{X},M) =\sum_ {x_i \in \hat{X}} K(x_i|M(c_{j_i}))$$, where $M(c_{j_i})=c_j^*$ is the object that the model $M$ assigns to the class $y_i$, and the class prediction for a new object $x$ is defined as: 
$$\hat{y} = c_j.{\arg\min}_{c_j^* \in C^*} K(x|c_j^*).$$ 
In short, we assign to each object the \textit{closest class} according to its algorithmic distance to one of the centroids in the set of objects $C^*$.
\end{defn}

Now, in a strong algorithmic sense, we can say that a classifier is optimal if the class assigned to each object fully describes this object minus incidental or incompressible information. In other words, if a classifier is optimal and we know the class of an object, then we know all its characteristics, except those that are unique to the object and not shared by other objects within the class. Formally, a classifier $f:X \rightarrow \{ c_1,\dots c_j \dots c_m\}$ is optimal with a degree of sophistication (in the sense of Koppel, \cite{koppel1991almost, koppel1991learning, antunes2009sophistication}) of $c$ if and only if, for every $x_i$, for any program $p$ and object $r$ such that $p(r)=x_i$ and $|p|+|r| \leq K(x_i)+c$, then $K(x_i|f(x_i)) \leq |r|+c$.
The next theorem shows that minimizing the stated cost function guarantees that the classifier is optimal in a strong algorithmic sense:

\begin{thm}\label{optimal}
If a classifier $f$ minimizes the cost function $J_a(X,f)$, then it is an optimal classifier,
\begin{proof}
Assume that $f$ is not optimal. Then there exist $x_i$ such that, for any class $c_j$, there exists a program $|p|$ and string $r$ such that $|p|+|r| \leq K(x_i)+c$, $p(r)=x_j$ and $K(x_i|c_j) > |r|+c$. Now, consider the classifier $f'$:

$$f'(x) = \begin{cases}
       f(x) & \text{if } x\neq x_i \\
       p & \text{otherwise}.
\end{cases}$$
It follows that
\begin{align*}
J(X,f')-J(X,f) &= K(x_i|p) - K(x_i|f(x_i)=c_j) \\
&>  K(x_i|p) - |r|+c\\
&\geq  |r| + O(1) - |r|+c  \geq c.
\end{align*}
\end{proof}

\end{thm}


\section{Approximating the Algorithmic Similarity Function}\label{condBDMSec}
While theoretically sound, the proposed algorithmic loss (Def.~\ref{aLoss}) and classification (Def.~\ref{clas}) cost functions rely on the uncomputable mathematical object $K$ (\cite{kolmo, chaitin3}). However, recent research and the existence of increasing computing power (\cite{d5, bdm}) have made available a number of techniques for the computable approximation of the non-conditional version of $K$. In this Section we present three methods for approximating the conditional algorithmic information function $K(x|y)$.

\subsection{Conditional CTM and Domain Specific CTM}\label{SCTM}

The Coding Theorem Method (\cite{d5}) is a numerical approximation to the algorithmic complexity of single objects. A generalization of CTM for approximating the conditional algorithmic complexity is the following:
\begin{defn}\label{condCTM}
Let $M: x \rightarrow y$ be a computable \textit{relation} and $P$ a finite set of pairs of the form $(y,x)$ corresponding to $M$. We define the \emph{conditional CTM} (with respect to $P$ and $M$) as: 

$$CTM(x|y) = \log_2 \left( \sum_{(y,x) \in P} \frac{1}{|P|}\right),$$, where $|P|$ is the cardinality of $P$. When $M$ is not a \textit{Turing complete} space, or $P$ does not contain an exhaustive computation of all possible pairs for the space, we say that we have a \emph{domain specific} CTM function.
\end{defn}

The previous Def.~is based on the Coding theorem (\cite{levin1974laws}), which establishes a relationship between the information complexity of an object and its algorithmic probability. If the relation $M$ approaches the space of all Turing machines, and $P$ is the set of all possible inputs and outputs for these machines, then at the limit, we have it that $CTM(x|y)=K(x|y)$. In the computable cases where we take a reduced (finite) set of Turing machines, we then have a (lower bounded) approximation to $K$. 

In the case where $x$ is the empty string and $M$ is the relation induced by the space of \textit{small} Turing machines with 2 symbols and 5 states, with $P$ computed exhaustively, $CTM$ approximates $K(x)$, and has been used to compute an approximation to the algorithmic complexity of small binary strings of up to size 12 (\cite{plos,d5}). Similarly, the space of small bi-dimensional Turing machines has been used to approximate the algorithmic complexity of square matrices of size up to $4\times 4$ (\cite{zenil2015two}).

When $M$ refers to a non-(necessarily) Turing complete space, or a computable input/output object different from ordered Turing machines, or if $P$ has not been computed exhaustively over this space, then we have a domain specific version of CTM, and its application depends on this space. This version of CTM is also an approximation to $K$, given that, if $M$ is a computable relation, then we can define a Turing machine with input $x$ that outputs $y$. However, we cannot guarantee that this approximation is consistent or (relatively) unbiased. Therefore we cannot say that it is domain independent.

At the time of writing this article, a database $P$ for conditional CTM over small Turing machines had yet to be computed. 

\subsection{Coarse Conditional BDM}

The Block Decomposition Method (BDM, \cite{bdm}) decomposes an object () into smaller parts for which there exist, thanks to CTM (\cite{bdm}),  \textit{good approximations} to their algorithmic complexity, and we then aggregate these quantities by following the rules of algorithmic information theory. We can apply the same concept to computing a \textit{coarse} approximation to the conditional algorithmic information complexity between two objects. Formally:
\begin{defn}\label{conditionalBDM}
We define the \emph{coarse conditional BDM} of $X$ with respect to the tensor $Y$ with respect to $\{\alpha_i\}$ as
 $$BDM(X|Y)=\sum_{(r_i,n_i) \in  Adj(X)-Adj(Y)}( CTM(r_i)+\log (n_i) )+ \sum_{Adj(X) \cap Adj(Y)} f(n^x_j,n^y_j)$$
\noindent{}where $\{\alpha_i\}$ is a partition strategy of the objects into smaller objects for which CTM values are known, $Adj(X)$ is the result of this partition for $X$ and $Y$ respectively, $n^x_j$ and $n^y_j$ are the multiplicity of the objects $r_j$ within $X$ and $Y$ respectively, and $f$ is the function defined as $$f(n^x_j,n^y_j)= \begin{cases} 0 & \mbox{if }  n^x_j=n^y_j\\ \log (n^x_j) & \mbox{otherwise.} \end{cases}$$ The sub-objects $r_i$ are called \emph{base objects} or \emph{base tensors} (when the object is a tensor) and are objects for which the algorithmic information complexity can be satisfactorily approximated by means of CTM.
\end{defn}

The motivation behind this definition is to enable us to consider partitions for the tensors $X$ and $Y$ into sets of subtensors $\langle x_i \rangle$ and $\langle y_i \rangle$, and then approximate the algorithmic information within the tensor $X$ that is not \textit{explained} by $Y$ by considering the subtensors $x_i$ which are not present in the partition $\langle y_i \rangle$. In other words, if we assume knowledge of $Y$ and its corresponding partition, then in order to describe the elements of the decomposition of $X$ using the partition strategy $\{\alpha_i\}$, we only need descriptions of the subtensors that are not $Y$. In the case of common subtensors, if the multiplicity is the same then we can assume that $X$ does not contain additional information, but that it does if the multiplicity differs.

The term $\sum_{Adj(X) \cap Adj(Y)} f(n^x_j,n^y_j)$ quantifies the additional information contained within $X$ when the multiplicity of the sub-tensors differs between $X$ and $Y$. This term is important in cases where such multiplicity dominates the complexity of the objects, cases that can present themselves when the objects resulting from partition are considerably smaller than the main tensors.

\subsection{Strong Conditional BDM}
The previous definition featured the adjective \textit{coarse} because we can define a stronger version of conditional BDM approximating $K$ with greater accuracy that uses conditional CTM. As explained in Section~\ref{coarseness}, one of the main \textit{weaknesses} of \textit{coarse} conditional BDM was the inability to detect the algorithmic relationship between \textit{base blocks}. This is in contrast with conditional CTM.

\begin{defn}\label{SConditionalBDM}

The \emph{strong conditional BDM of $X$ with respect to $Y$ corresponding to the partition strategy $\{\alpha_i\}$} is 

$$BDM(X|Y) = \displaystyle\min_P \sum_{((r^x_i,n^x_i), (r^y_i,n^y_i)) \in P}( CTM(r^x_i|r^y_j)+f(n^x_i,n^y_i))$$ where $P$ is a  \textit{pairing} of the base elements in the decomposition of $X$ and $Y$ where the elements of $X$ can appear only once in a pair but without restrictions as to the elements of $Y$. This is a functional relation $(r^x_i,n^x_i), \mapsto (r^y_i,n^y_i)$, and $f$ is the same function as specified in Def.~\ref{conditionalBDM}. If conditional CTM is used, then we say that we have \emph{strong conditional CTM}. 
\end{defn}

While we assert that the \textit{pairing strategy} minimizing the given sum will yield the best approximation to $K$ in all cases, prior knowledge of the algorithmic structure of the objects can be used to facilitate the computation by reducing the number of possible pairings to be explored, especially when using the \textit{domain specific} version of conditional BDM. For instance, if two objects are known to be produced from local dynamics, then restricting the algorithmic comparisons by considering pairs based on their respective position on the tensors will, with high probability, yield the best approximation to their algorithmic distance.

\subsubsection{The Relationship Between Coarse and Strong BDM}
It is easy to see that under the same partition strategy, strong conditional BDM will always present a better approximation to $K$ than its coarse counterpart. If the partition of two tensors $\langle r_i^x\rangle$ and $\langle r_i^x\rangle$ does not share an algorithmic connection other than subtensor equality, i.e. there exists $r^x_i = r^y_j$, then this is the best case for coarse BDM, and applying both functions will yield the same approximation to $K$. However, if there exist two base blocks where 
$CTM(r^x_i|r^y_j)< CTM(r^x_i)$ then $K(X|Y) - O(\log^2 A)\leq \text{strong }BDM(X|Y) < \text{coarse }BDM(X|Y)$ where $A$ is the \textit{diminishing} error incurred in proposition 1 in \cite{bdm}.
Moreover, unlike coarse conditional BDM, the accuracy of the approximation offered with the strong variant will improve in proportion to the size of the base objects, ultimately converging towards CTM and then $K$ itself.

The properties of strong and coarse conditional BDM and their relation with entropy are shown in the appendix (Section \ref{app}). In particular, we show that conditional BDM is \textit{well behaved} by defining \textit{joint} and \textit{mutual BDM} (Section~\ref{jointAndMutual}), and we show that its behavior is analogous to the corresponding Shannon's entropy functions. We also discuss the relation that both measures have with entropy (Section \ref{coarseness}), showing that, in the worst case, we converge towards conditional entropy.

\section{Algorithmic Optimization Methodology}

In the previous sections we proposed algorithmic loss and cost functions (Def.~\ref{aLoss}) for supervised learning tasks, along with means to compute approximations to these theoretical mathematical objects. Here we ask how to perform model parameter optimization based on such measures. Many of the most widely used optimization techniques rely on the cost function being (\textit{sufficiently}) \textit{differentiable, smooth and convex} (\cite{armijo1966minimization}), for instance gradient descent and associated methods (\cite{bottou2010large,adam}). In the next section we will show that such methods are not adequate for the algorithmic cost function.

\subsection{The Non-smooth Nature of the Algorithmic Space}\label{nonSmooth}
Let us start with a simple bilinear regression problem. Let 
\begin{equation}\label{f}
f(a,b)= 0.1010101\dots \times a + 0.01010\dots \times b 
\end{equation} be a linear function used to produce a set of 20 random data points $\hat{X}$ of the form $(a,b, f(a,b))$, and $M (a,b)=s_1 \times a + s_2 \times b$ be a proposed model whose parameters $s_1$ and $s_2$ must be optimized in order to, hopefully, fit the given data.

\begin{figure}[ht]
  \centering
  \subfloat{{\includegraphics[width=7.5cm]{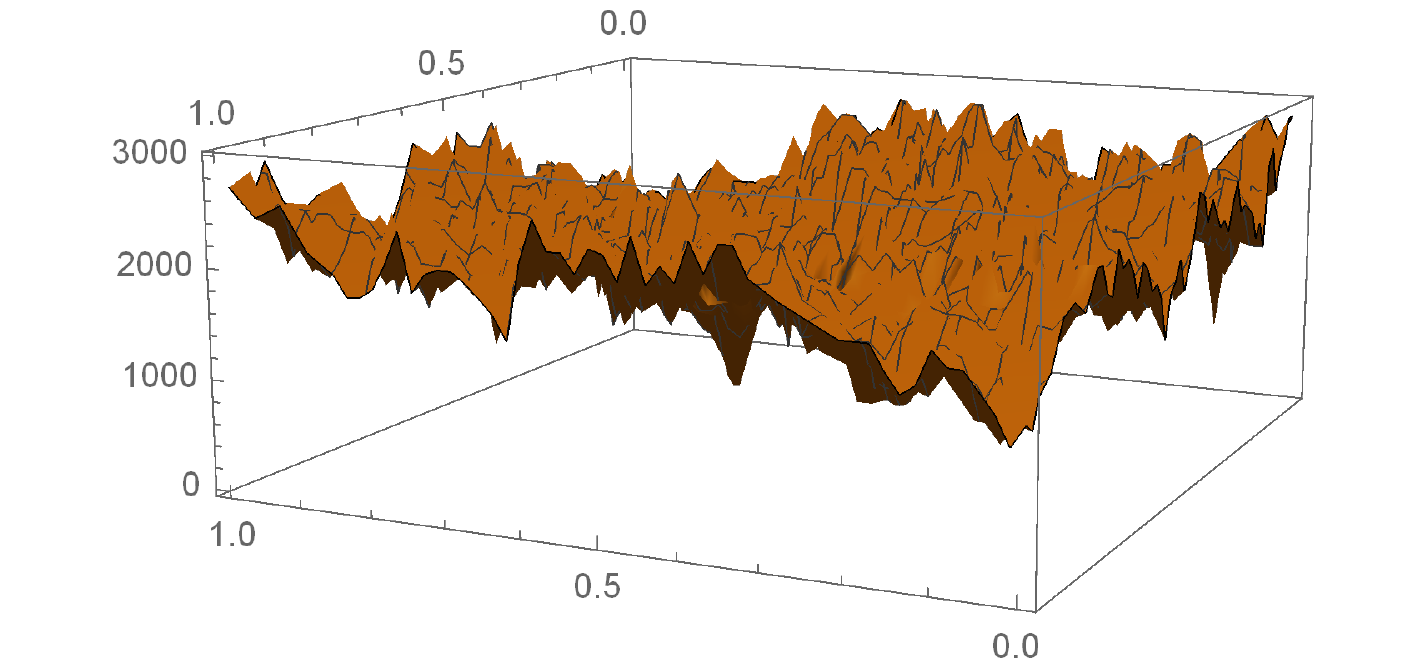} }}%
  \qquad
  \qquad
    \subfloat{{\includegraphics[width=7cm]{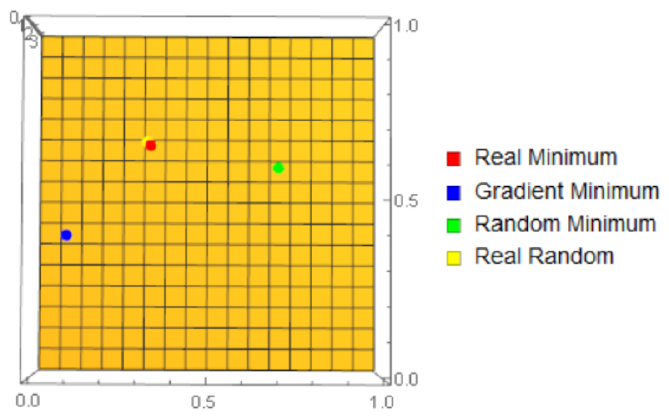} }}%
  \caption{On the left we have a visualization of the algorithmic cost function, as approximated by coarse BDM, corresponding to a simple bilinear regression problem. From the plot we can see the complex nature of the optimization problem. On the right we have confirmation of these intuitions in the fact that the best performing optimization algorithm is a random pooling of 5000 points.}
  \label{noCon}
\end{figure}

According to the Def.~\ref{aLoss}, the loss function associated with this optimization problem is $J(\hat{X},M) = \sum_{(a,b,y) \in \hat{X}}K(y|M(a,b))^2.$ A visualization of the surface resulting from this function, where $K$ was approximated by coarse conditional BDM (Def.~\ref{conditionalBDM}) with a partition of size 3 can be seen on the left of Figure~\ref{noCon}. From the plot we observe that the resulting curve is not smooth and that gradient based approaches would fail to converge toward a non-local minimum. This observation was evaluated by applying several optimization techniques: gradient descent (constrained to a square of radius 0.25 around the solution), random search, and a \textit{purely} random search. The purely random algorithm simply pooled 5000 random points and chose the point where the cost function evaluated was the lowest. At the right of the Fig.~\ref{noCon} we can see that this random pooling of points yielded the \textit{optimization technique}. It is well understood that a random pooling optimization method like the one we performed is not scalable to larger, more complex problems. However, the function $f$ has an algorithmic property that will allow us to construct a more efficient optimization method, that we will call \textit{algorithmic parameter optimization}.

\subsection{Algorithmic Parameter Optimization}

The established link between algorithmic information and algorithmic probability theory (\cite{levin1974laws}) provides a path for defining optimal (under the only assumption of computable algorithms) optimization methods.  The central question in the area of parameter optimization is the following: Given a data set $\hat{X}=\langle x, y\rangle$, what is the best model $M$ that satisfies $M(x)=y$, and hopefully, will extend to pairs of the \textit{same phenomena} that are not present in $\hat{X}$? 

Algorithmic probability theory establishes and quantifies the fact that the \textit{most probable computable program is also the least complex one}~\cite{solo,levin}, thereby formalizing a principal of parsimony such as Ockham's razor. Formally, we define the \emph{algorithmic parameter optimization problem} as the problem of finding a model $M$ such that \textit{(a) minimizes} $K(M)$ and \textit{(b) minimizes the cost function} $$J_a(\hat{X},M) = \sum_{(x,y) \in \hat{X}}K(y|M(x))^2 .$$

By searching for the solution using the \textit{algorithmic order} we can meet both requirements in an \textit{efficient amount of time}. We start with the least complex solution, therefore the most probable one, and then we move towards the most complex candidates, stopping once we find a \textit{good enough} value for $J_a$ or after a determined number of \textit{steps}, in the manner of other optimizations methods.

\begin{defn}\label{algOp}
Let $M$ be a model,  $J_a$ the algorithmic cost function, $\hat{X}$  the training set and finally let  $\Sigma=\{\sigma_1, \sigma_2, \dots,\sigma_i,\dots\}$ be the \textit{parameter space} which is ordered according to their \textit{algorithmic order} (from least to most algorithmically complex). Then the \textit{simple algorithmic parameter optimization (or algorithmic search)} is performed by

\begin{algorithm}[ht]\label{prog}
 $ \text{minCost} =\infty$\;
 \While{\textit{condition}}{$i$++\;
  \If{$J_a(\hat{M},\sigma_i) < \emph{minCost}$}{
   $\text{minCost} =J_a(\hat{M},\sigma_i )$\;
   param = $\sigma_i$\;
   }
 }
 \textit{Return }$\sigma_i$ 
\end{algorithm}

\noindent{}where the halting \textit{\textit{condition}} is defined in terms of the number of iterations or a specific value for $J_a$.
\end{defn}

The algorithmic cost function is not expected to reach zero. In a \textit{perfect fit} scenario, the loss of a sample is the relative algorithmic complexity of $y$ with respect to the model itself, which can be unbounded. Depending on the learning task, we can search for heuristics to define an approximation to the optimum value for $J_a$, and thus end the search when a close enough optimization has been reached, resembling the way in which the number of clusters is naturally estimated with algorithmic information itself~\cite{nmi}, stopping the process when the model's complexity starts to increase rather than decrease when given the data as conditional variable. Given the semi-uncomputable nature of $K$, there is no general method to find such conditions, but they can be approximated. Another way to define a stopping condition is by combining other \textit{cost} functions, such as the MSE or accuracy over a validation set in the case of classification. What justifies Def.~\ref{algOp} is the importance of the ordering of the parameter space and the expected execution time of the program provided. 

By Def.~\ref{algOp}, it follows that the parameter space is countable and computable. This is justified, given that any program is bound by the same requirements. For instance, in order to fit the output of the function $f$ (Eq. \ref{f}) by means of the model $M$, we must optimize over two continuous parameters $s_1$ and $s_2$. Therefore the space of parameters is composed of the pairs of real numbers $\sigma_i=[\sigma^i_1\;\sigma^i_2]$. However, a computer cannot fully represent a real number, using instead an approximation by means of a fixed number of bits. Since this second space is finite, so is the parameter space and the search space which is composed of pairs of binary strings of finite size, the algorithmic information complexity value of which can be approximated by BDM or CTM. Furthermore, as the next two examples will show, for algorithmic optimization a search space based on binary strings can be considered an asset that can be exploited to speed up the algorithm and improve performance, rather than a hindrance because of its lower accuracy in representing continuous spaces. This is because algorithmic search is specifically designed to work within a computable space.

Now, consider a fixed model structure $M$. Given that the algorithmic parameter optimization always finds the \textit{lowest algorithmically complex parameters} that fit the data $\hat{X}$ within the halting condition, the resulting model is the most algorithmically plausible model that meets the restrictions imposed by the Def.~of $M$. This property results in a natural tendency to avoid over-fitting. Furthermore, algorithmic optimization will always converge significantly more slowly to overly complex models that will tend to over-fit the data even if they offer a \textit{better explanation} of a reduced data set $\hat{X}$. Conversely, algorithmic parameter optimization will naturally be a poor performer when inferring models of high algorithmic complexity. Finally, note that the method can be applied to any cost function, preserving the above properties. Interestingly, this can potentially be used as a method of regularization in itself.

\subsubsection{On the Expected Optimization Time}

Given the way that algorithmic parameter optimization works, the optimization time, as measured by the number of iterations, will converge faster if the optimal parameters have low algorithmic complexity. Therefore they are more plausible in the algorithmic sense. In other words, if we assume that, for the model we are defining, the parameters have an underlying algorithmic cause, then they will be found faster by algorithmic search, sometimes much faster. How much faster depends on the problem and its algorithmic complexity. In the context of artificial evolution and genetic algorithms, it has been previously shown that, by using an algorithmic probability distribution, the exponential random search can be sped up to quadratic (\cite{chaitin:EvolofMutaSoft, ChaitinBook, hernandez2018algorithmically}).  

Following the example of inferring the function in section \ref{nonSmooth}, the mean and median BDM value for the parameter space of pairs of 8-bit binary strings are 47.6737 and 47.7527, respectively; while the optimum parameters $\{0.10101011, 0.01010101\}$ have a BDM of 44.2564. This lower BDM value confirms the intuition that binary representations of both parameters have an algorithmic source (repeating 10 or 01). The difference in value might seem small on a continuum, but in algorithmic terms it translates into an exponential absolute distance between candidate strings: the optimum parameters are expected to be found at least $2^{3.4}$ times faster by algorithmic optimization (compared to a search within the space). The optimum solution occupies position 1026 out of 65,281 pairs of strings. Therefore the optimum for this optimization problem can be found within 1026 iterations, or nearly 65 times faster.

The assumption that the optimum parameters have an underlying simplicity bias is strong, but has been investigated~\cite{kamal,zenilchaitin} and is compatible with principles of parsimony. This bias favours objects of interest that are of low algorithmic complexity, though they may appear random, For example, the decimal expansions of the constant $\pi$ or $e$ to an accuracy of 32 bits have a BDM value of 666.155 and 674.258, respectively, while the expected BDM for a random binary string of the same size is significantly larger:  $\approx 681.2$. This means that we can expect them to be found significantly faster, according to algorithmic probability--- about $2^8$ and $2^7$ time steps faster, respectively, compared to a random string--- by using algorithmic optimization methods.

At the same time, we are aware that, for the example given, mathematical analysis-based optimization techniques have a perfect and efficient solution in terms of the gradient of the MSE cost function. While algorithmic search is faster than random search for a certain class of problems, it may be slower for another large class of problems. However, algorithmic parameter optimization (Def.~\ref{algOp}) is a domain and problem-independent general method. While this new field of algorithmic machine learning that we are introducing is at an early stage of development. in the next sections we set forth some further developments that may help boost the performance of our algorithmic search for specific cases, such as greedy search over the subtensors, and there is no reason to believe that more boosting techniques will not be developed and introduced in the future.

\section{Methods}

\subsection{Traversing Non-smooth Algorithmic Surfaces for Solving Ordinary Differential Equations}\label{ODESec}

Thus far we have provided the mathematical foundation for machine learning based on the power of algorithmic probability at the price of operating on a non-smooth loss surface in the space of algorithmic complexity. While the directed search technique we have formulated succeeds with discrete problems, here we ask whether our tools generalize to problems in a continuous domain. To gain insight, we evaluate whether we can estimate parameters for ordinary differential equations. Parameter identification is well-known to be a challenging problem in general, and in particular for continuous models when the data-set is small and in the presence of noise. 
Following \cite{dua2011simultaneous}, as a sample system we have $\frac{dz_1}{dt}=-\theta_1 z_1$ and $\frac{dz_2}{dt}=-\theta_1 z_1 - \theta_2 Z_2$ (Eq. 2) with hidden parameters $[\theta_1 \; \theta_2]=[5 \; 1]$ and $z(t = 0)=[1 \; 0]$. Let $Ev(t,[\theta_1 \; \theta_2])$ be a function that \textit{correctly} approximates the numerical value corresponding to the given parameters and $t$ for the ODE system.
Let us consider a \textit{model} composed of a binary representation of the pair $[\theta_1 \; \theta_2]$ by a 16 bit string where the first 8 bits represent $\theta_1$, the last 8 bits $\theta_2$ for a parameter search space of size $2^{16}=65\,536$, and where within these 8 bits the first 4 represent the integer part and the last 4 the fractional part. Thus the hidden solution is represented by the binary string `0101000000010000'.

\subsection{Finding Computable Generative Mechanisms}\label{fECA}

\begin{figure}[ht]
  \centering
  \subfloat{{\includegraphics[width=0.3\textwidth]{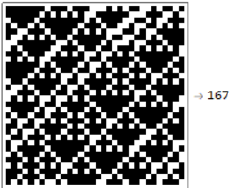} }}%
  \qquad
  \qquad
    \subfloat{{\includegraphics[width=0.3\textwidth]{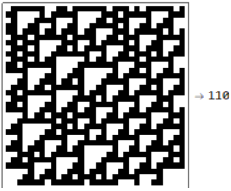}}}%
  \caption{Two $32 \times 32$ images and their respective classes. The images represent the evolution of the automata 167 and 110 for two different randomly generated 32-bits binary strings.}
  \label{samples}
\end{figure}

An elementary cellular automaton (ECA) is a discrete and linear binary dynamical system where the state of a node is defined by the states of the node itself and its two adjacent neighbours (\cite{nks}). Despite their simplicity, the dynamics of these systems can achieve Turing completeness. The task was to classify a set of $32 \times 32$ black and white images representing the evolution of one of eleven elementary cellular automata according to a random 32-bit binary initialization string. The automata were $$C=\{167, 11, 129, 215, 88, 32, 237, 156, 173, 236, 110\}.$$ Aside from the Turing-complete rule with number 110, the others were randomly selected among all 256 possible ECA. The \texttt{training set} was composed of 275 black and white images, 25 for each automaton or `\textit{class}'. An independent \texttt{validation set} of the same size was also generated, along with a \texttt{test-set} with 1\,375 evenly distributed samples. An example of the data in these data sets is shown in figure~\ref{samples}.

First we will illustrate the difficulty of the problem by training neural networks with simple topologies over the data. In total we trained three \textit{naive}\footnote{We say that a NN topology is \emph{naive} when its design does not use specific knowledge of the target data.} neural networks that consisted of a flattened layer, followed by either 1, 2, 3 or 4 fully connected linear layers, ending with a softmax layer for classification. The networks were trained using ADAM optimization for 500 rounds. Of these 4 models, the network with 3 linear layers performed best, with an accuracy of 40.3\%.

However, as shown in~\cite{celClass}, it is possible to design a deep network that achieves a higher accuracy for this particular classification task. This topology consists of 16 convolutional layers with a kernel of $2 \times 3$, which was specifically chosen to fit the rules of ECA, a pooling layer that aggregates the data of all the convolutions into a vector of one dimension of length 16, and 11 linear layers (256 in the original version) connected to a final softmax unit. This network achieves an accuracy of 98.8\% on the \texttt{test set} and 99.63\% on the \texttt{training set} after 700 rounds. This specialized topology is an example of how, by using prior knowledge of the algorithmic organization of the data, it is possible to guide the variance of a neural network towards the algorithmic structure of the data and avoid overfitting. However, as we will show over the following experiments, this is a specialized ad-hoc solution that does not generalize to other tasks.

\subsubsection{ Algorithmic-probability Classifier Based on Coarse conditional BDM}\label{first}

The algorithmic probability model chosen consists of eleven $16\times 16$ binary matrices, each corresponding to a sample class, denoted by $M$, encompassing members $m_i \in M$. \textit{Training} this model, using Def.~\ref{clas}, the loss function $$\sum_ {x_i \in \texttt{test set}} K(x_i|M(y_i)),$$ is minimized, where $M(y_i)=m_j$ is the object that the model assigns to the class $y_i$. Here we approximate the conditional algorithmic complexity function $K$ with the \textit{coarse conditional BDM} function, then proceed with algorithmic information optimization over the space of the possible $16 \times 16$ binary matrices in order to minimize the computable cost function $J(\texttt{test set},M) =\sum_ {x_i \in \texttt{test set}} BDM(x_i|M(y_i)).$  However, an elementary cellular automaton can potentially use all the 32-bits of information contained in a binary initialization string, and the algorithmic difference between each cellular automaton is bounded and relatively small (within 8-bits). Furthermore, each automaton and initialization string was randomly chosen without regard to an algorithmic cause. Therefore we cannot expect a significant speed-up by using an algorithmic search, and it would be nearly equivalent to randomly searching through the space of $16 \times 16$ matrices, which will take a length of time of the order of $O(2^{256})$. Nevertheless, we perform a \textit{greedy block optimization} version of algorithmic information optimization:\\

\begin{figure}
     \subfloat[]{%
       \includegraphics[width=0.18\textwidth]{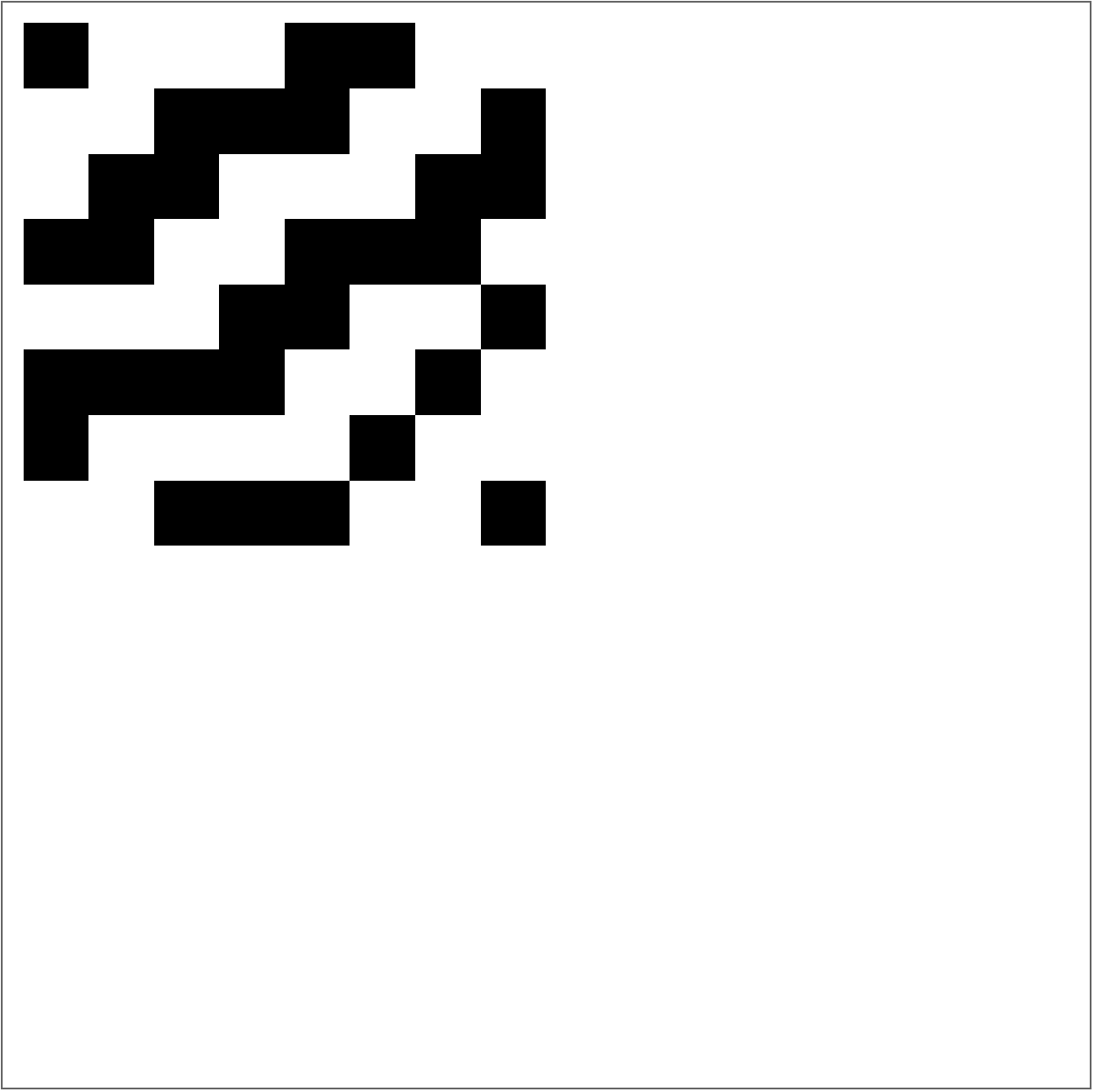}
     }
     \hfill
     \subfloat[]{%
       \includegraphics[width=0.18\textwidth]{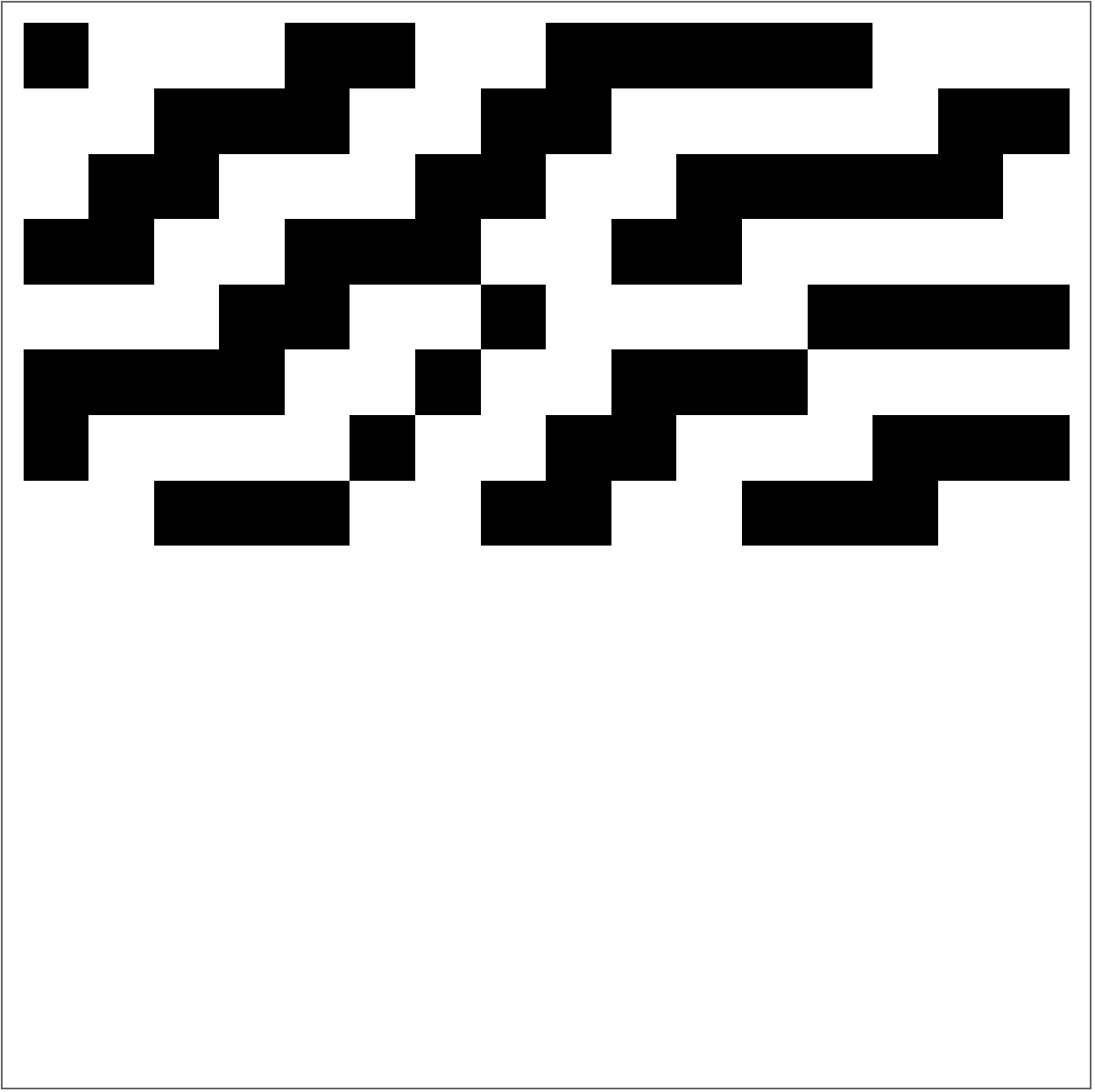}
     }
     \hfill
     \subfloat[]{%
       \includegraphics[width=0.18\textwidth]{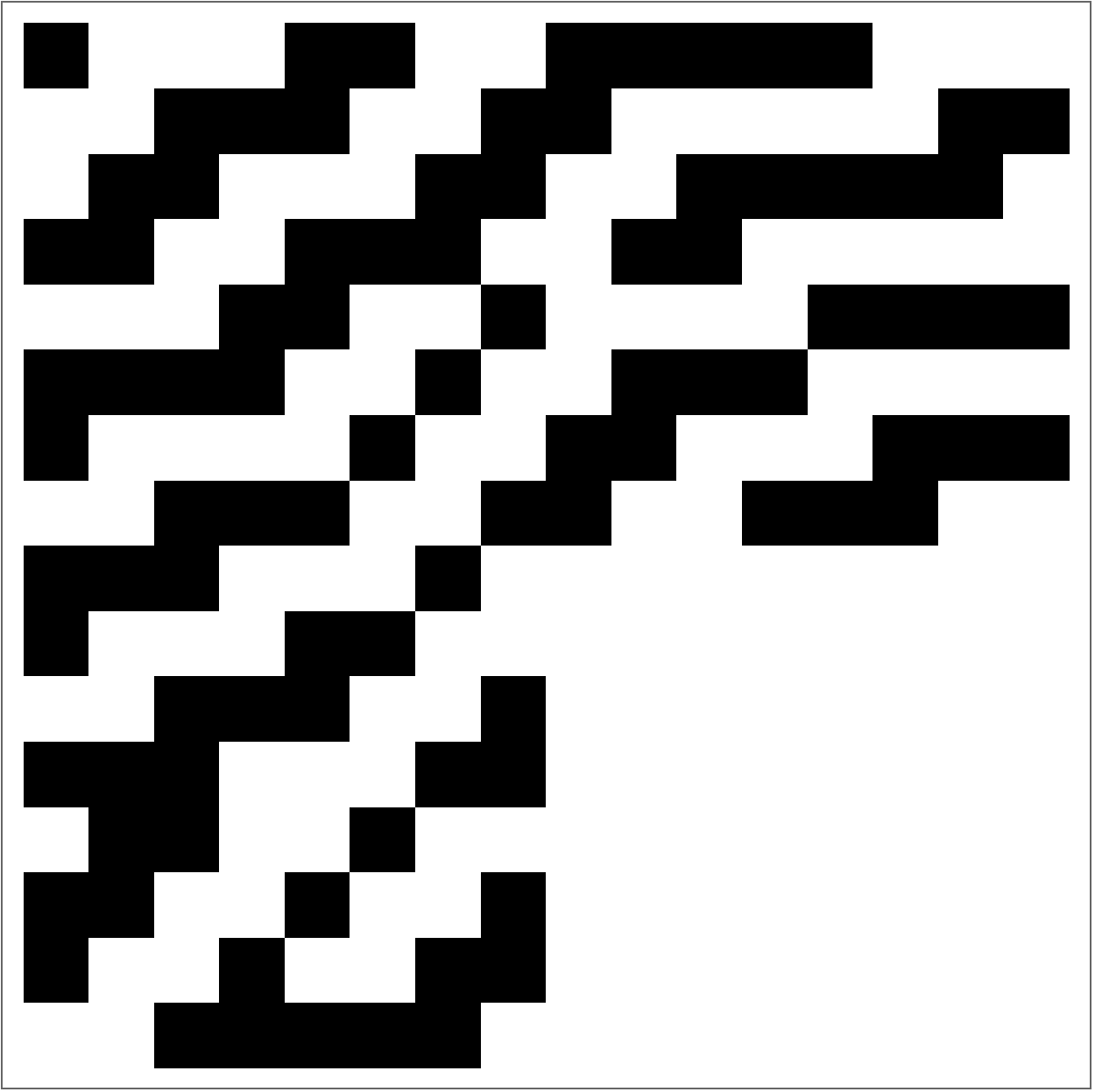}
     }
     \hfill
     \subfloat[]{%
       \includegraphics[width=0.18\textwidth]{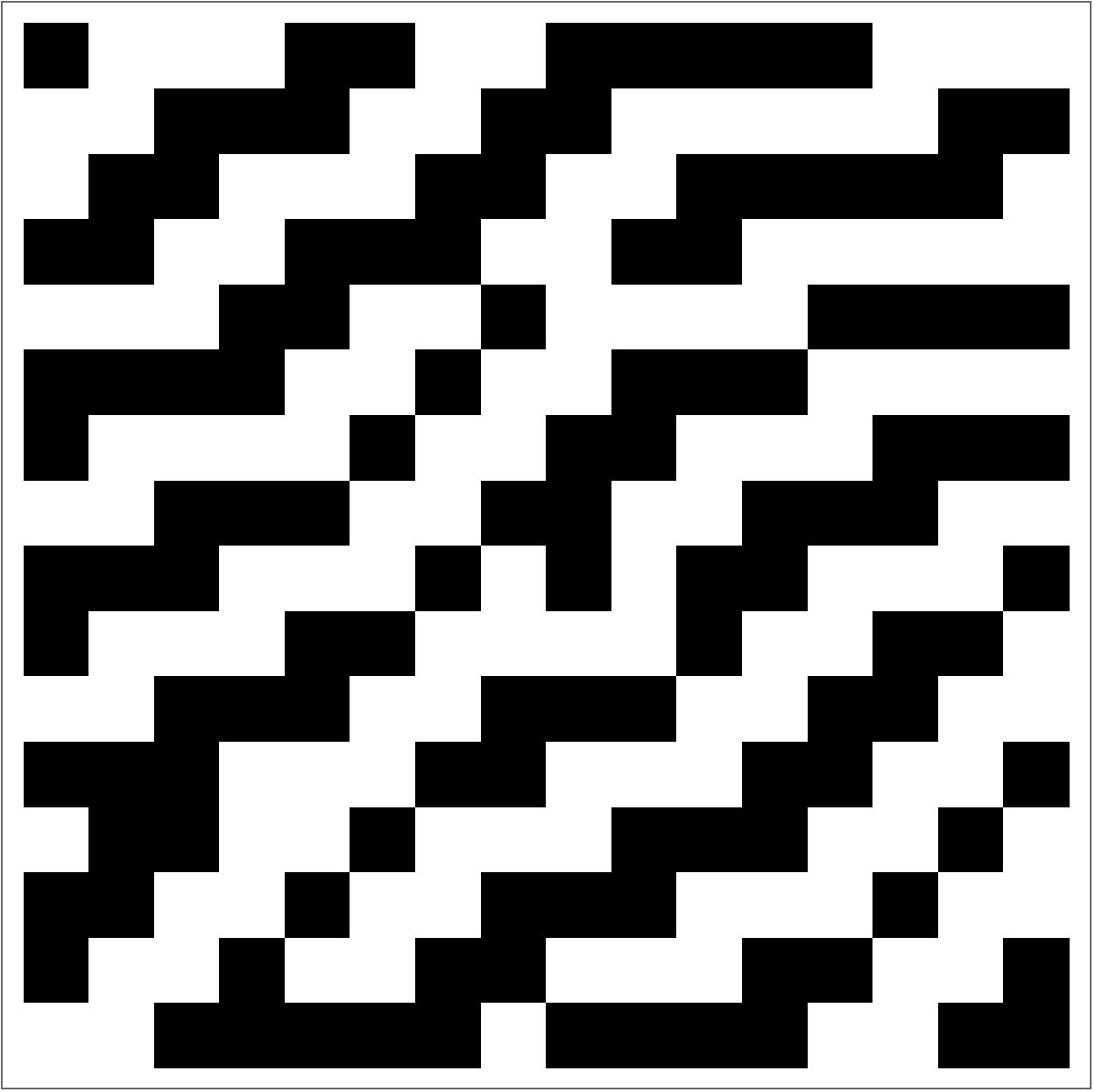}
     }
     \caption{The evolution of the centre for the four steps of the greedy algorithmic information optimization method used to train the model in the first experiment. This classifier centre corresponds to class 11. }
     \label{steps}
  \end{figure}

\begin{myitemize}
\item First, we start with the eleven $16 \times 16$ matrix of 0s.
\item Then, we perform algorithmic optimization, but only changing the bits contained in the upper left $8 \times 8$ submatrix. This step is equivalent to changing all the $2^{8 \times 8}$ bits in the quadrant, searching for the matrix that minimizes the cost function.
\item After minimizing with respect to only the upper left quadrant, we minimize over the upper right $8 \times 8$ quadrant.
\item We repeat the procedure for the lower left and lower right quadrants.
\end{myitemize}
\noindent{}These four steps are illustrated in Fig.~\ref{steps}for the class corresponding to the automaton $11$.

\subsection{Finding the Initial Conditions for Cellular Automata} \label{fI}

The next problem was to classify black and white images representing the evolution of elementary cellular automata. In this case, we are classifying according to the \textit{initialization string} that produced the corresponding evolution \textit{for a randomly chosen automaton}. The \textit{classes} for the experiment consisted of 10 randomly chosen binary strings, each 12 bits in length. These strings correspond to the binary representation of the following integers:
$$704, 3572, 3067, 3184, 1939, 2386, 2896, 205, 828, 3935.$$ The \texttt{training}, \texttt{validation} and \texttt{test sets} each consisted of two hundred $12 \times 4$ binary images. These images represent the evolution to \textit{4 steps} of one of the 10 strings within the first 128 cellular automata rules (to avoid some trivially symmetric cases) by means of a randomly chosen cellular automaton. It is important to note that the first row of the evolution (the initialization) was removed. Otherwise this classification task would be trivial.

We trained and tested a group of neural network topologies on the data in order to establish the difficulty of the classification task. These networks were an (adapted version of) Fernandes' topology and 4 \textit{naive} neural networks that consisted of a flattened (fully-connected) layer, followed by 1, 2, and 5 groups of layers,
each consisting of a fully connected linear layer with rectilinear activation (ReLU) function followed by a dropout layer, ending with a linear layer and a softmax unit for classification.  The adaptation of the Fernandes topology was only for the purpose of changing the kernel of the pooling layer to $2\times 9$ to take into account the non-square shape of the data. All networks were trained using the ADAM optimizer.

The best performing network was the shallower one, which consists of a flattened layer, followed by a fully connected ReLU, a dropout layer, a linear layer with 10 inputs and a sotfmax unit. This neural network achieved an accuracy of 60.1\%. 
At 18.5\%, the performance of Fernandes' topology was very low, being barely above random choice. This last result is to be expected, given that the topology is \textit{domain specific}, and should not be expected to extend well to different problems, even though at first glance the problem may seem to be related.

\subsubsection{ Algorithmic-probability Classifier Based on Strong conditional BDM}\label{second}

The algorithmic probability model $M$ chosen for these tasks consisted of eleven 12-bit binary vectors. The model was trained using \textit{algorithmic information greedy block optimization} by first optimizing the loss function over the 6 leftmost bits and then over the remaining six.

However, for this particular problem, the \textit{coarse} version of \textit{conditional BDM} proved inadequate for approximating the universal algorithmic distance $K(x_i|M(y_i))$, for which reason we opted to use the stronger version.  For the \textit{stronger} version of conditional BDM we approximated the local algorithmic distance $CTM(x|s)$, where $x$ is a binary matrix of size $6 \times 4$ and $s$ is a binary vector of length 6, in the following way.

\begin{myitemize}
\item First, we computed all the outputs of all possible 12-bit binary strings for each of the first 128 ECA for a total of 528,384 pairs of 12 bit binary vectors and $12 \times 4$ binary matrices, forming the set of pairs $(s, x) \in P$.
\item Then, by considering only the inner 6 bits of the vectors (discarding the 3 bits on the left and the 3 bits on the right) and, similarly, the inner $6 \times 4$ submatrix, we defined $$CTM(x|s) = \log_2 \left( \sum_{(s,x) \in P} \frac{1}{|P|}\right),$$ where $|P|$ is the cardinality of $P$. This \textit{cropping} was done to solve the frontier issue of finite space ECA. 
\item If a particular pair $(s,x)$ was not present in the database, then considering that $-\log_2(1/528,384)= 19.01$, we defined $CTM(x|s)=20$. This means that the algorithmic complexity of $x$ given $s$ is at least 20 bits.
\item In the end we obtained a database of the algorithmic distance between all \textit{6} bit vectors and their respective \textit{$6 \times 4$} possible outputs.
\end{myitemize}

The previous procedure might at first seem to be too computationally costly. However, just as with Turing Machine based CTM (\cite{soler2014calculating, bdm}), this computation only needs to be done once, with the data obtained being reusable
in various applications.

The trained model $M$ consisted of 10 binary vectors that, as expected, corresponded to the binary expansion of each of the classes. The accuracy of the classifier was 95.5\% on the \texttt{test set}.

\subsection{Classifying NK Networks}\label{fourth}

An NK network is a dynamical system that consists of a binary Boolean network where the parameter $n$ specifies the number of nodes or vertices and $k$ defines the number of incoming connections that each vertex has (\cite{dubrova2005kauffman, kauffman1969metabolic, aldana2003boolean}). Each node has an associated $k$-ary Boolean function which uses the states of the nodes corresponding to the incoming connections to determine the new state of the nodes over a discrete time scale. The number of incoming connections $k$ defines the stability (or lack thereof) of the network.

Given the extensive computational resources it would require to compute a CTM database, such as the one used in section \ref{fourth}, for Boolean networks of 24 nodes we opted to do a classification based only on the algorithmic complexity of the samples as approximated by BDM. This approach is justified, considering that according to the definition \ref{clas}, an algorithmic information model for the classifier can consist of three sets. Each of these sets is composed of all possible algorithmic relations, including the adjacency matrix and related binary operations, corresponding to the number of incoming connections per node (the parameter $k$). Therefore, given the combinatorial order of growth of these sets, we can expect the quantity of information required to specify the members of each class to increase as a function of $k$.

Specifically, the number of possible Boolean operations of degree $k$ is $2^{2^k}$ and the number of possible adjacency matrices is $n\times \binom nk$. It follows that the total number of possible network topologies is $n^2 \times 2^{2^k} \times  \binom nk$, and the expected number of bits required to specify a member of this set is $\log(n^2 \times 2^{2^k} \times  \binom nk)$. Therefore, the expected algorithmic complexity of the members of each class increases with $k$ and $n$. With $n$ fixed at 24 we can do a \textit{coarse} algorithmic classification simply according to the algorithmic complexity of the samples, as approximated by BDM.

Following this idea we defined a classifier where the model $M$ consisted of the mean BDM value for each of the classes in the \texttt{training set}
$ M = \{1\rightarrow 2671.46 ,2\rightarrow 4937.35, 3\rightarrow 6837.64 \}.$
The prediction function measures the BDM of the sample and assigns it to the class \textit{centre} that is the closest. This classifier achieved an accuracy of 71\%. Alternatively, we employed a nearest neighbour classifier using the BDM values of the \texttt{training set}, which yielded virtually identical results. For completeness sake, we recreated the last classifier using entropy to approximate the algorithmic information theory $K$. The accuracy of this classifier was 37.66\%.


For classifying according to the \textit{Boolean rules} assigned to each node,
we used 10 randomly generated (ordered) lists of 4 binary Boolean rules. These rules were randomly chosen (with repetitions) from $And$, $Or$, $Nand$ and $XOr$, with the only limitation being that $Nand$  had to be among the rules. Since the initial state for the network was the vector $\{0,0,0,0\}$, at least one XOr was needed in order to generate an evolution other than forty 0s. Then, to generate the samples,
each list of binary rules was associated with a number of random topologies (with $k$=2). The \texttt{training} and \texttt{validation sets} were composed of 20 samples for each class (200 samples in total) while the \texttt{test set} contained 2000 samples.

To classify according to network \textit{topology} we used 10 randomly generated topologies consisting of 10 binary matrices of size $10 \times 10$,
which represented the adjacency matrices of the chosen topologies. The random matrices had the limitation that each column had to contain two and only two 1s,
so the number of incoming connections corresponds to $k=2$. Then, to generate a sample we associated one of the chosen topologies with a random list of rules. This list of rules was, again, randomly chosen from the same 4 Boolean functions and with the limitation that $XOr$ had to be a member. The \texttt{training} and \texttt{validation sets} were composed of 20 samples for each class (200 samples in total) while the \texttt{test set} contained 2000 samples.

\subsection{Classifying Kauffman networks}

Kauffman networks are a special case of Boolean NK networks where the number of incoming connections for each node is two, that is, $k=2$ (\cite{atlan1981random}). This distinction is important because when $k$ = 2 we have a \textit{critical point} that ``\textit{exhibits self-organized critical behaviour}''; below that ($k=1$) we have too much regularity, a (\textit{frozen state}) and beyond it ($K \geq 3$) we have \textit{chaos}.

\subsubsection{Algorithmic-probability Classifier based on conditional CTM}

For this problem we used a different type of \textit{algorithmic cluster centre}. For the \textit{Boolean rules} classifier, the model $M$ consisted of ten lists of Boolean operators. More precisely, the model consisted of binary strings that codified a list composed of each of the four Boolean functions used ($And$, $Or$, $Nand$ and $XOr$) as encoded by the \texttt{Wolfram Language}. For the \textit{topology} classifier, the model consisted of 10 binary matrices representing the possible network topologies. 

The use of different kinds of models for this task showcases another layer of abstraction that can be used within the wider framework of algorithmic probability classification: \textit{context}. Rather than using binary tensors, we can use a structured object that has meaning for the underlying problem. Yet,
as we will show next, the underlying mechanics will remain virtually unchanged.  

Let's go back to the definition \ref{clas}, which states that to train both models we have to minimize the cost function $$J(\hat{X},M)=\sum_ {x_i \in \texttt{test set}} K(x_i|M(y_i)).$$ So far we have approximated $K$ by means of conditional BDM. However, given that at the time of writing this article a sufficiently complete conditional CTM database has yet to be computed, we have estimated the CTM function by using instances of the computable objects, as previously shown in section \ref{second}. Moreover, owing to the non-local nature of NK networks and the abstraction layer that the models themselves are working at, rather than using BDM, we have opted to use a context dependent version of CTM directly. In the last task we will show that BDM can be used to classify a similar, yet more general problem.

Following similar steps to the ones used in section \ref{second}, by computing all the 331,776 possible NK networks with $n=4$ and $k=2$, we compiled two lists of pairs $P_1$ and $P_2$ that contained, respectively, the pairs $(t,x)$ and $(r,x)$, where $t$ is the topology and $r$ is the list of rules that generated the 40-bit vector $x$, which represents the evolution to ten steps of the respective networks. Next, we defined the $CTM(x|s)$ as:

$$CTM(x|s) = \log_2 \left( \sum_{(s,x) \in P_i} \frac{1}{|P_i|}\right),$$ or as 19 if the pair is not present on either of the lists. Then we approximated $K$ by using the defined $CTM$ function directly.

\subsection{Hybrid Machine Learning}

So far we have presented supervised learning techniques that, in a way, diverge. In this section we will introduce one of the ways in which the two paradigms can coexist and complement each other, combining statistical machine learning with an algorithmic-probability approach.

\subsubsection{Algorithmic Information Regularization}

The choice of an appropriate level of model complexity that avoids both under- and over-fitting is a key hyperparameter in machine learning. Indeed, on the one hand, if the model is too complex, it will fit the data used to construct the model very well but generalize poorly to unseen data. On the other hand, if the complexity is too low, the model will not capture all the information in the data. This is often referred to as the bias-variance trade-off, because a complex model will exhibit large variance, while an overly simple one will be strongly biased. Most traditional methods feature this choice in the form of a free hyperparameter via, e.g., what is known as regularization.

A family of mathematical techniques or processes that has been developed to control over-fitting of a model goes under the rubric 'regularization', which can be summarized as the introduction of \textit{information} from the model to the training process in order to prevent over-fitting of the data. A widely used method is the Tikhonov regularization (\cite{Tikhonov,reg}), also known as ridge regression or weight decay, which consists in adding a penalty term to the cost function of a model, which increases in direct proportion to the norms of the variables of the model. This method of regularization can be formalized in the following way:  Let $J$ be the cost function associated with the model $M$ trained over the data set $\hat x$, $p$ a \textit{model weighting function} of the form $p: M \mapsto \mu$, where $\mu \in \mathbb{R}^+$, and $\lambda$ a positive real number. The (hyper)parameter $\lambda$ is called a \textit{regularization parameter}; the product $\lambda p(M)$ is known as the \textit{regularization term} and the \textit{regulated cost function $J'$} is defined as 
\begin{equation}\label{reg}
J'(\hat x,M,\lambda)=J(\hat x,M)+\lambda p(M).
\end{equation}
The core premise of the previous function is that we are disincentivizing fitting towards certain parameters of the model by assigning them a higher cost in proportion to $\lambda$, which is a hyperparameter that is learned empirically from the data.  In current machine learning processes, the most commonly used weighting functions are the sum of the $L_1,L_2$ norms of the linear coefficients of the model, such as in ridge regressions (\cite{ridge}).

We can employ the basic form of equation \ref{reg} and define a regularization term based on the algorithmic complexity of the model and, in that way, disincentivize training towards algorithmically complex models, thus increasing their algorithmic plausibility. Formally:
\begin{defn} \label{alReg}
Let $J$ be the cost function associated with the model $M$ trained over the data set  $\hat x$, $K$ the universal algorithmic complexity function, and $\lambda$ a positive real number. We define the \textit{algorithmic regularization} as the function $$J_K(\hat x,M,\lambda)=J(\hat x,M)+\lambda K(M).$$
\end{defn}
The justification of the previous definition follows from algorithmic probability and the coding theorem: Assuming an underlying computable structure, the most probable model that fits the data is the least complex one. Given the universality of algorithmic probability, we argue that the stated definition is general enough to improve the plausibility of the model of any machine learning algorithm with an associated cost function. Furthermore, the stated definition is compatible with other regularization schemes.

Just as with the algorithmic loss function (Def.~\ref{lossS}), the resulting function is not smooth, and therefore cannot be optimized by means of gradient-based methods. One option for minimizing this class of functions is by means of algorithmic parameter optimization (Def~\ref{algOp}). It is important to recall that computing approximations to the algorithmic probability and complexity of objects is a recent development, and we hope to promote the development of more powerful techniques.

\subsection{Algorithmic-probability Weighting}

Another, perhaps more direct way to introduce algorithmic probability into the current field of machine learning, is the following. Given that in the field of machine learning all model inference methods must be computable, the following inequality holds for any fixed training methodology:
\begin{equation}\label{eq1}
K(M^*) \leq K(\hat X )+K(M^I)+O(1),
\end{equation}
\noindent{}where $M^*$ is the fitted model, $\hat X$ is the training data, $M^I$ is the model with the parameters during its initialization and $O(1)$ corresponds to the length of the program implementing the training procedure. Now, using common initialization conventions, $M^I$ either has very low algorithmic complexity or very high (it's random), in order to not induce a bias in the model. Thus the only parameter on the right side of the inequality that can be optimized is  $K(\hat X)$.  It follows that increasing the algorithmic plausibility of a model can be achieved by reducing the algorithmic complexity of training set $\hat X$, which can be achieved by preprocessing the data and \textit{weighting} each sample using its algorithmic information content, thus optimizing in the direction of samples with lower algorithmic complexity.

Accordingly, the heuristic for our definition of algorithmic probability weighting is that, to each training sample, we assign an importance factor (weight) according to its algorithmic complexity value, in order to increase or diminish the loss incurred by the sample. Formally:

\begin{defn}\label{weight}
Let  $J$ be a cost function of the form $$J(\hat X,M)=g(L(y_1,y'_1),\dots, L(y_i,y'_i),\dots,L(y_n,y'_n)).$$We define the \emph{weighted approximation to the algorithmic complexity regularization of $J$} or \emph{algorithmic probability weighting} as $$J_{w,k}(\hat X,M,f) = g(f(K(x_1))  \cdot L(y_1,y'_1),\dots, f(K(x_i)) \cdot L(y_i,y'_i),\dots, f(K(x_n))  \cdot L(y_n,y'_n)),$$ where $f(K(x_i))$ is a function that \textit{weights} the algorithmic complexity of each sample of the training data set in \textit{a way that is constructive with respect to the goals of the model}.
\end{defn}
We have opted for flexibility regarding the specification of the function $f$. However taking into account the noncontinuous nature of $K$, we have recommended a discrete definition for $f$. The following characterization has worked well with our trials and we hold that it is general enough to be used in a wide number of application domains:

\begin{equation*}
f(x_i,X, \langle \gamma_k \rangle, \langle \varphi_k \rangle)=
\begin{cases}
\gamma_1 &\text{if } K(x_i) \in Q(\varphi_1, K(C(x_i))) \\
\gamma_2 &\text{if }  K(x_i) \in Q(\varphi_2, K(C(x_i)))\\
\dots\\
\gamma_k &\text{if } K(x_i) \in Q(\varphi_k,K(C(x_i)))\\
\dots\\
\gamma_j &\text{if } K(x_i) \in Q(\varphi_j, K(C(x_i)))
\end{cases}
\end{equation*}

where $\gamma_k$ and $\varphi_k$ are hyperparameters and $K(x_i) \in Q(\varphi_k, K(C(x_i)))$ denotes that $K(x_i)$ belongs to the $\varphi_k$-ith quantile of the distribution of algorithmic complexities of all the samples belonging to the same class as $x_i$.

As its names implies, the previous Def.~\ref{weight} can be considered analogous to sample weighting, which is normally used as a means to confer predominance or diminished importance on certain samples in the data set according to specific statistical criteria, such as \textit{survey weights} and \textit{inverse variance weight}~\cite{hartung}. However, a key difference of our definition is that \textit{traditional} weighting strategies rely on statistical methods to infer values from the population, while with algorithmic probability weighting we use the universal distribution for this purpose. This makes algorithmic probability weighting a \textit{natural extension or universal generalization} of the concept of sample weighting, and given its universality, it is domain independent.

Now, given that the output of $f$ and its parameters are constant from the \textit{point of view} of the parameters of the model $M$, it is easy to see that if the original cost function $J$ is continuous, differentiable, convex and smooth, so is the weighted version $J_{w,k}$. Furthermore, the stated definition is compatible with other regularization techniques, including other weighting techniques, while the algorithmic complexity of the samples can be computed by numerical approximation methods such as the Block Decomposition Method.

\section{Results}

\subsection{Estimating ODE parameters}

\begin{figure}[ht]
  \centering
\includegraphics[width=8cm]{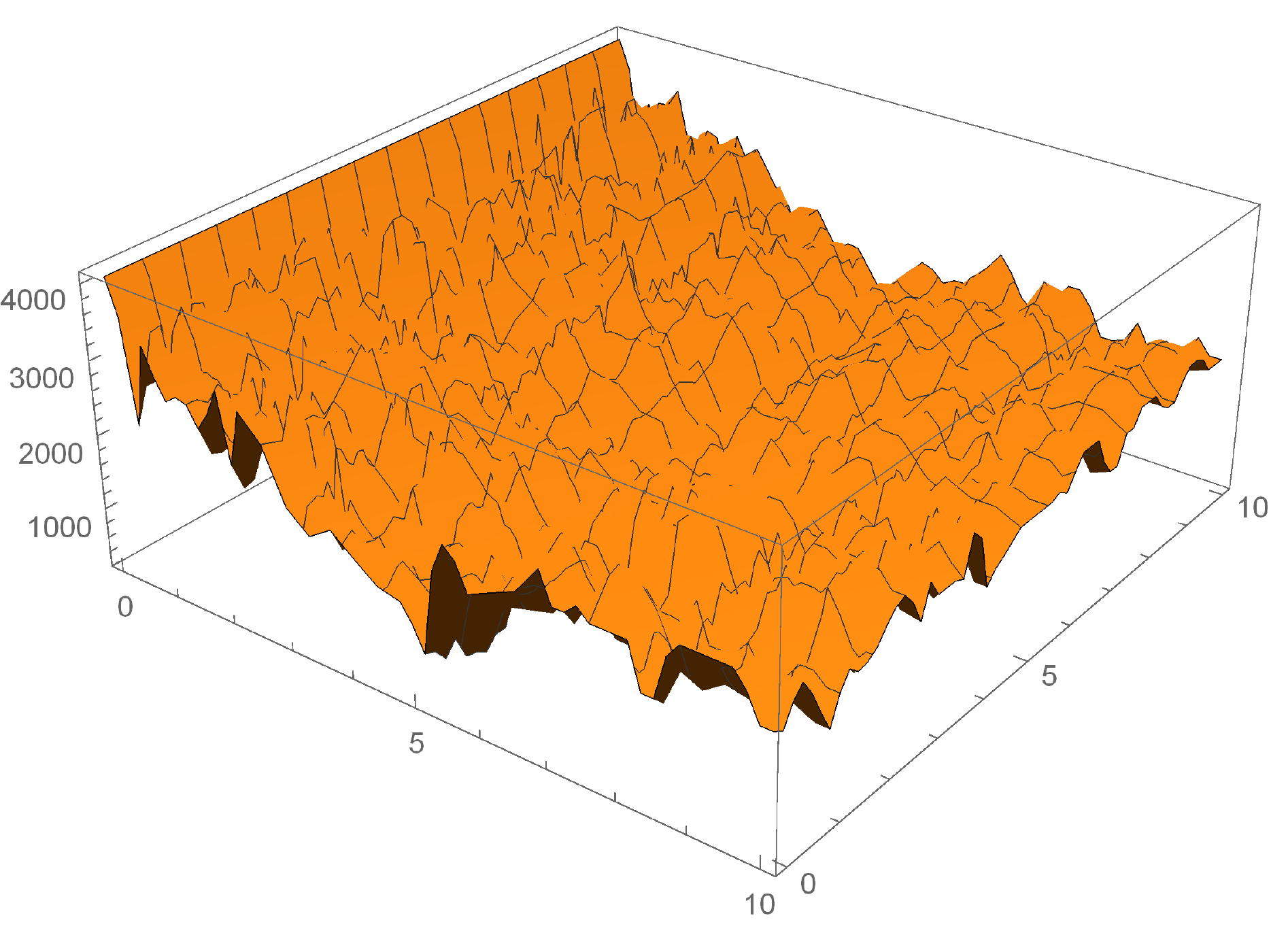}
  \caption{A visualization of the algorithmic cost function, as approximated by coarse BDM, corresponding to the parameter approximation of an ordinary differential Eq. 2. From the surface obtained we can see the complexity of finding the optimal parameters for this problem. Gradient based methods are not optimal for optimizing algorithmic information based functions.}
  \label{ODES}
\end{figure}

\begin{figure}
  \centering
\includegraphics[width=8cm]{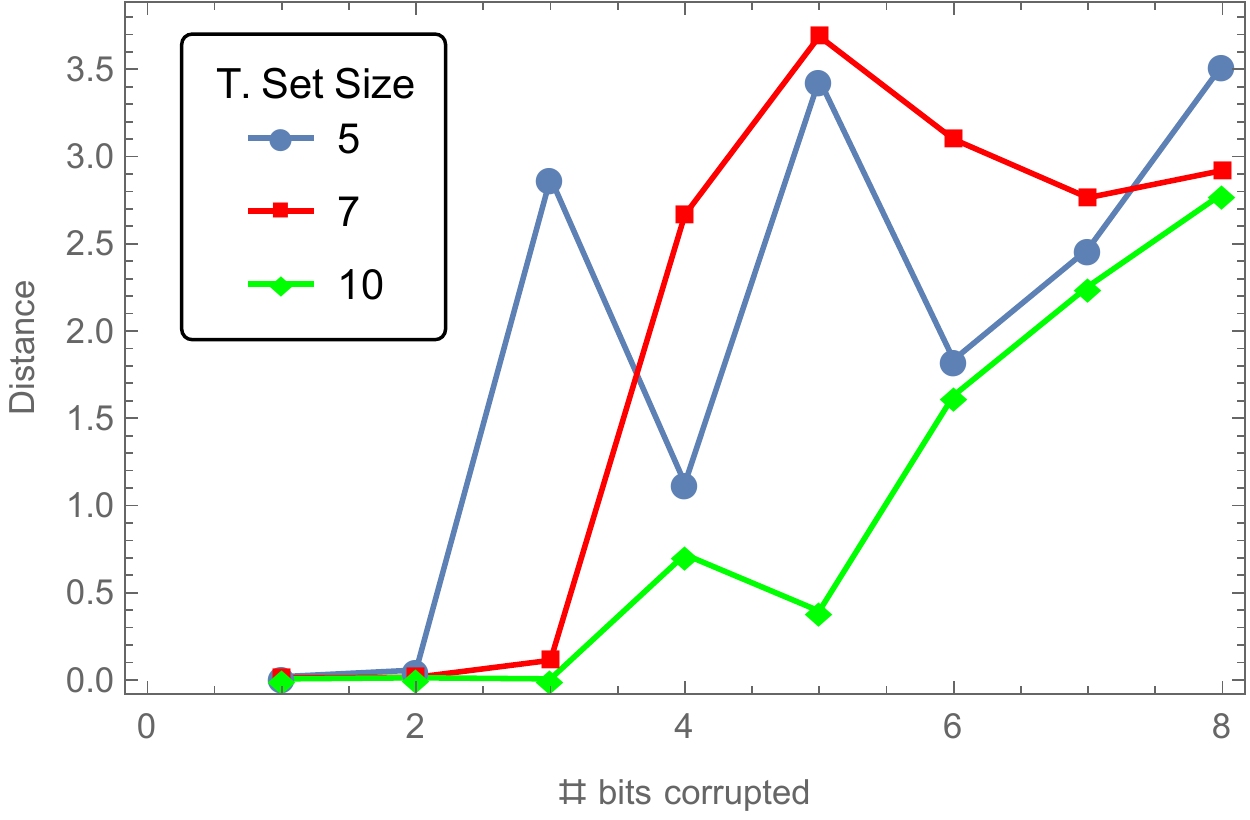}
  \caption{The average Euclidean distance between the solution inferred by algorithmic optimization and the hidden parameters of the ODE in Section~\ref{ODESec} when a number of bits of the binary representation of the labelled data has been randomly \textit{corrupted} (flipped), from 1 to 8 bits. The binary representation of the states $z_1$ and $z_2$ has an accuracy of 8 bits each, or 16 bits for the pair. At the maximum corruption level, 50\% of the bits that represent a pair of outputs in the sample set are being flipped, destroying any kind of information within each sample. The average was taken from ten random sample sets of size 5, 7, and 10. Each set was computed by randomly selecting times $t$ between 0 and 1 in intervals of size 0.05, then corrupting the corresponding output by the specified number of bits ($y$ axis). From the  results it can be seen that algorithmic optimization is resilient up to corruptions of 4 to 5 bits, or $> 25\%$ percent of the data, even when using relatively small training data sets.}
  \label{Noise}
\end{figure}

A key step to enabling progress between a fundamentally discrete theory such as computability and algorithmic probability, and a fundamentally continuous theory such as that of differential equations and dynamical systems, is to find ways to combine both worlds. As shown in Section~\ref{nonSmooth}, optimizing the parameters with respect to the algorithmic cost function is a challenge (Fig.~\ref{ODES}). Following algorithmic optimization, we note that parameters (5 and 1) have low algorithmic complexity due to their functional relation. This is confirmed by BDM, which assigns the unknown solution to the ODE a value of 153.719 when the expected complexity is approximately 162.658, which means that the number of more complex parameter candidates than $[\theta_1 \; \theta_2]=[5 \; 1]$ must be on the order of $2^{8}$. Within the parameter space, the solution is at the position 5\,093 out of 65\,536. Therefore the exact ODE solution can be found within less than 6 thousand iterations following the simple algorithmic parameter optimization (Def.~\ref{algOp}) by consulting the function $Ev$. Furthermore, for the training set of size 10 composed of the pairs $z_1(t),z_2(t)$ corresponding to the list $$t={0.1,0.2,\dots,0.8,0.9,1.0},$$, we need only 2 samples to identify the solution, supporting the expectation that algorithmic parameter optimization ensures a solution with high probability, despite a low number of samples \textit{as long as the solution has low complexity} in a relatively low number of iterations. This is proof-of-principle that our search technique can not only be used to identify parameters for an ODE problem, but also affords the advantage of faster convergence (fewer iterations), requiring less data to solve the parameter identification problem. In Fig. \ref{Noise}, equivalent to the pixel attacks for discrete objects, we show that the parameter identification is robust to even more than 25\% of additive noise. Operating in a low complexity regime--- as above---is compatible with a principal of parsimony such as Ockham's razor, which is empirically found to be able to explain data simplicity bias~\cite{algo,kamal,zenilbadillo}, suggesting that the best explanation is the simplest, but also that what is modelled is not algorithmically random~\cite{zenilchaitin}.

\subsection{Finding Generative Rules of ECA}

Following optimization, a classification function was defined to assign a new object to the class corresponding to the centre $m_j$ to which it is the closest according to the algorithmic distance $BDM(x|m_j)$. The classifier obtained reaches an accuracy of 98.1\% on the test set and of 99.27\% on the training set (table \ref{resTables1}).

\begin{table}[ht]
\vspace{-.5cm}
  \caption{The accuracy of the Tested Classifiers}
  \label{resTables1}
  \centering
  \begin{tabular}{ccc}
    \toprule
    Classifier    & \texttt{Test Set}  & \texttt{Training Set}  \\
    \midrule
     \multicolumn{1}{c}{Naive Networks}                   \\
    \cmidrule(r){1-3}
    1 & 38.88\% & 95.63\% \\
    2 & 39.70\% & 95.63\\
    3 & 40.36\% & 100\%\\
    4 & 39.05\% & 100\%\\
    \cmidrule(r){1-3}
    Fernandes' & 98.8\% & 99.63\%\\
    \cmidrule(r){1-3}
    Algorithmic Class. & 98.1\% & 99.27\%\\
    \bottomrule
  \end{tabular}
\end{table}

From the data we can see that the algorithmic classifier outperformed the four \textit{naive} (or simple) neural networks, but it was outperformed slightly by the Fernandes classifier, built expressly for the purpose.
But as we will show over the following sections, this last classifier is less robust and is domain specific. 

Last but not least, we have tested the \textit{robustness} of the classifiers by measuring how good they are at resisting \textit{one-pixel attacks} (\cite{su2019one}). A one-pixel attack occurs when a classifier can be fooled into misclassifying an object by changing just a small portion of its information (one pixel). Intuitively, such small changes should not affect the classification of the object in most cases, yet it has recently been shown that deep neural network classifiers present just such vulnerabilities. 

Algorithmic information theory tells us that algorithmic probability classifier models should have a relatively high degree of resilience in the face of such attacks: if an object belongs to a class according to a classifier it means that it is algorithmically \textit{close} to a centre defining that class. A one-pixel attack constitutes a relatively small information change in an object. Therefore there is a relatively high probability that a one-pixel attack would not alter the information content of an image enough to increase the distance to the centre in a significant way. 
In order to test this hypothesis, we systematically and exhaustively searched for vulnerabilities in the following way:
\texttt{(a)}One by one, we flipped (from 0 to 1 or vice versa) each of the $32 \times 32$ pixels of the samples contained in the \texttt{test data}.
\texttt{(b)} If a flip was enough to change the assigned classification for the sample, then it was counted as a vulnerability.
\texttt{(c)}Finally, we divided the total number of vulnerabilities found by the total number of samples in order to obtain an \textit{expected number of vulnerabilities} per sample. The results obtained are shown in Table~{ref}.

\begin{table}

  \caption{Expected Number of Vulnerability Per Sample}
 \label{robTable}
  \begin{center}
  \begin{tabular}{ccccc}
    \toprule
    Classifier    & Total Vulnerabilities & Per Sample & Percentage of Pixels  \\
    \midrule
    Fernandes' (DNN) & 190,850 & 138.88 & 13.56\% \\
    Algorithmic Classifier & 15,125 & 11 & 1\% \\
    \bottomrule
  \end{tabular}
  \end{center}
\end{table}
  
From the results we can see that for the DNN, 13.56\% of the pixels are vulnerable to one-pixel attacks, and that only 1\% of the pixels manifest that vulnerability for the algorithmic classifier. These results confirm our hypothesis that the algorithmic classifier is significantly more robust in the face of \textit{small} perturbations compared to the deep network classifier designed without a specific purpose in mind. It is important to clarify that we are not stating that it is not possible to increase the robustness of a neural network, but rather pointing out that algorithmic classification has a high degree of robustness \textit{naturally}.

\subsection{Finding Initial Conditions}

The accuracy obtained using the different classifiers is represented in Table~\ref{resTables2}. Based on these results we can see that the algorithmic classifier performed significantly better than the neural networks tested. Furthermore, the first two \textit{naive} topologies have enough variance present to have a good fit vis-a-vis the training set, in an obvious case of over-fitting. The domain specific Fernandes topology maintained a considerably high error rate---exceeding 80\%---over 3,135 ADAM training rounds. 
It is important to note that in this case collisions, that is, two samples that belong to two different classes, can exist. Therefore it is
impossible to obtain 100\% \textit{perfect} accuracy. An exhaustive search classifier that searches through the space for the corresponding initialization string reached an accuracy of 97.75\% over the \texttt{test set}.\\
In order to test the generalization of the CTM database computed for this experiment, we tested our algorithmic classifying scheme on a different instance of the same basic premise: binary images of size $24 \times 4$ that correspond to the output of twenty randomly selected binary strings of 24 bits each for a randomly chosen ECA. The number of samples per class remains at 20 for the \texttt{training}, \texttt{validation} and \texttt{test sets}. The results are shown in the following table. For this case the algorithmic classifier increased its accuracy to 96.56\%. Thanks to the additional data, the neural networks also increased their accuracy to 64.11\% and 61.74\% for the first and second topology, respectively.

\subsubsection{Network Topology Algorithmic-information Classifier}\label{fifth}

The results are summarized in Table~\ref{resTables4}. Here we can see that only the \textit{coarse} BDM algorithmic information classifier---with 70\% accuracy---managed to reach an accuracy that is significantly better than random choice, validating our method. 

\begin{figure}
  \centering

  \subfloat{{\includegraphics[width=0.34\textwidth]{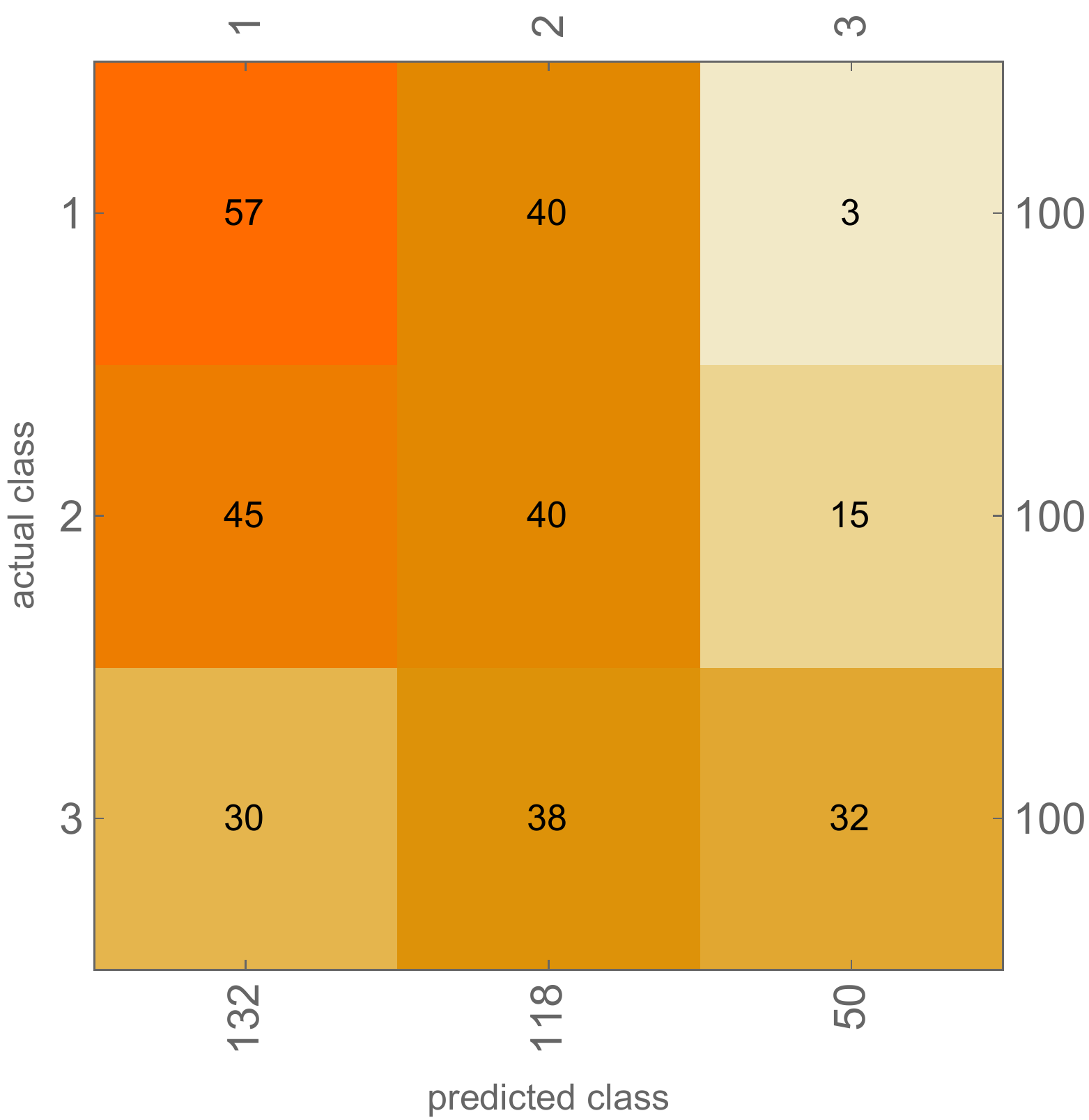} }}%
  \qquad
  \qquad
    \subfloat{{\includegraphics[width=0.34\textwidth]{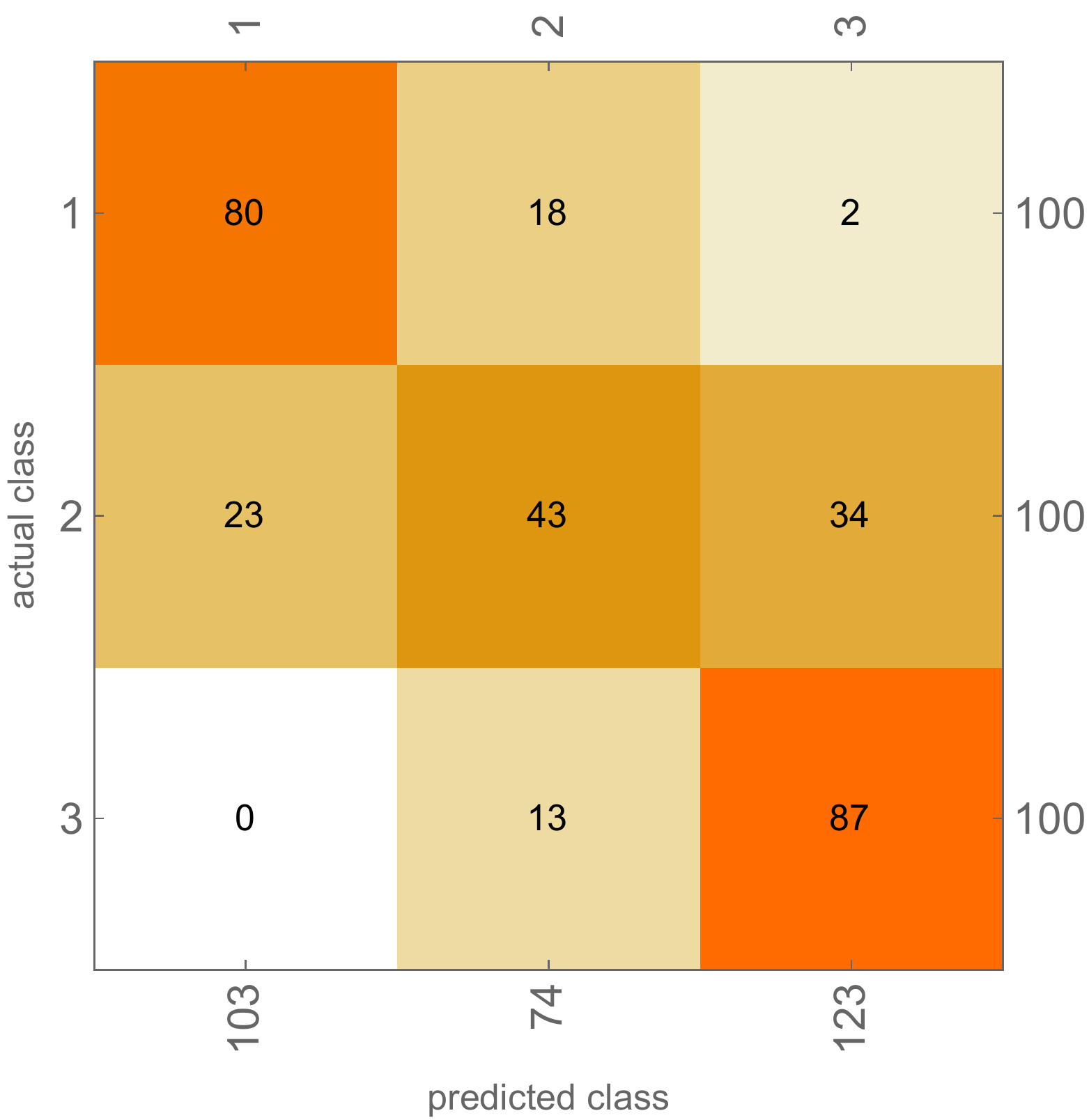}}}%
  \caption{The confusion matrix for the neural network (left) and the algorithmic information classifier (right) while classifying binary vectors representing the degree of connection (parameter $k$) of 300 randomly generated Boolean NK networks. From the plots we can see that the algorithmic classifier can predict with relatively high accuracy the elements belonging to class $k=1$ and $k=3$. The neural network used is considerably more random in its predictions.}
  \label{conPlots}
\end{figure}

Furthermore, by analyzing the confusion matrix plot (Figure~\ref{conPlots}) we can see that the algorithmic classifier performs (relatively) well at classifying the \textit{frozen} and \textit{chaotic} networks, while the deep learning classifier seems to be random in its predictions. The fact that the critical stage was harder to classify is evidence of its \textit{rich} dynamics, accounting for more varied algorithmic behaviours. 

A second task was to classify a set of binary vectors of size 40 that represent the evolution of an NK network of four nodes ($n=4$) and two incoming connections ($k=2$). Given that an NK network is defined by two causal features, the topology of the network and the Boolean function of each node, we divided the second task in two: classifying according to its topology and according to the underlying Boolean rules.

\subsection{Classifying Kauffman Networks}\label{task3}

The task was to determine whenever a random Boolean network belonged
to the frozen, critical or chaotic phase by determining when $k=1$, 2 or 3. Furthermore, we used the full range of possible unary, binary and tertiary Boolean operations corresponding to each of the functions associated with a node. The objects to classify were binary vectors of size 240 bits that represented the evolution of networks of 24 nodes to ten steps with incoming connections of degree 1, 2 or 3. The \texttt{training}, \texttt{validation} and \texttt{test} sets were all of size 300, with 100 corresponding to each class. 
For this more general, therefore harder classification task, we used larger objects and data sets. The objects to classify were binary vectors of size 240 bits that represented the evolution of networks of 24 nodes to ten steps with incoming connections of degree 1, 2 or 3. The \texttt{training}, \texttt{validation} and \texttt{test} sets were all of size 300, with 100 corresponding to each class. 

For the task at hand we trained the following classifiers: a neural network, gradient boosted trees and a convolutional neural network. The first neural network had a \textit{naive} classifier that consisted of a ReLU layer, followed by a Dropout layer, a linear layer and a final softmax unit for classification. For the convolutional model we used prior knowledge of the problem and used a specialized topology that consisted of 10 convolutional layers with a kernel of size 24, each kernel representing a stage of the evolution, with a ReLU, a pooling layer of kernel size 24, a flattened layer, a fully connected linear layer and a final softmax layer. The tree-based classifier manages an accuracy of 35\% on the \textit{test set}, while the naive and convolutional neural networks managed an accuracy of 43\% and 31\% percent respectively. Two of the three classifiers are nearly indistinguishable from random classification, while the naive neural network is barely above it.

For comparison purposes, we trained a neural network and a logistic regression classifier on the data. The neural network consisted of a \textit{naive} topology consisting of a ReLU layer, followed by a dropout layer, a linear layer and a softmax unit. The results are shown in Table~\ref{resTables3}. 

From the results obtained we can see that the neural network, with 92.50\% accuracy, performed slightly better than the algorithmic classifier (91.35\%) on the \texttt{test set}. The logistic regression accuracy is a bit further behind,
at 82.35\%.

However, the difference in the performance of the \texttt{topology test set} is much greater, with both the logistic regression and the neural network reaching very high error rates. In contrast, our algorithmic classifier reaches an accuracy of 72.4\%.

\subsection{A First Experiment and Proof of Concept of Algorithmic-probability Weighting}

As a first experiment in algorithmic weighing, we designed an experiment using the MNIST dataset of hand written digits \cite{mnist}. This dataset, which consists of a training set of 60,000 labelled images representing hand written digits from 0 to 9 and a test set with 10,000 examples, was chosen given its historical importance for the field and also because it offered a direct way to deploy the existing tools for measuring algorithmic complexity via \textit{binarization} without compromising the integrity of the data. 

The binarization was performed by using a simple mask: if the value of a (gray scale) pixel was above 0.5 then the value of the pixel was set to one, using zero in the other case. This transformation did not affect the performance of any of the deep learning models tested, including the LeNet-5 topology (\cite{lecun1998gradient}), in a significant way. 

\begin{figure}[ht!]
  \centering
  \subfloat{{\includegraphics[width=5cm]{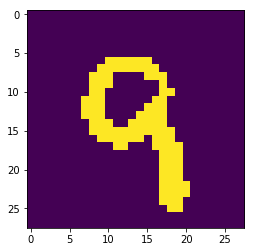} }}%
  \qquad
  \qquad
    \subfloat{{\includegraphics[width=5cm]{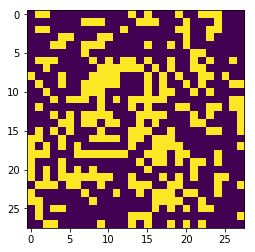} }}%
  \caption{At left we have an image representing a binarized version of a randomly chosen sample. At right we have the \textit{salted} version of the same sample, with 30\% of its pixels randomly shuffled. }
  \label{salting}
\end{figure}

Next we \textit{salted} or \textit{corrupted} 40\% of the training samples by randomly shuffling 30\% of their pixels. An example of these corrupted samples can be seen in Figure~\ref{salting}. With this second transformation we are reducing the useful information within a random selected subset of samples by a random amount, thus simulating a case where the expected amount of incidental information is high, as in the case of data loss or corruption.

Finally, we trained 10 neural networks with increasing depth, setting aside 20\% of the training data as a verification set, thereby obtaining neural networks of increasing depth and, more importantly, variance. The topology of these networks consisted of a flattened layer, followed by an increasing number of fully connected linear layers with rectified linear (ReLU) activation functions, and a final softmax layer for classification. In order to highlight the effect of our regularization proposal, we abstained from using other regularization techniques and large batch sizes. For instance, optimizing using variable learning rates such as RMSProp along with small stochastic batches is an alternative way of steering the samples away from the salted samples.

For purposes of comparison, the neural networks were trained with and without weighting, using the option \texttt{sample\_weight} for the \texttt{train\_on\_batch} on \texttt{Keras}.The training parameters for the networks, which were trained using \texttt{Keras} on \texttt{Python 3}, were the following:

\begin{itemize}
\item Stochastic gradient descent with batch size of 5\,000 samples.
\item 40 epochs, (therefore 80 \textit{training stages}), with the exception of the last model with 10 ReLU layers, which was trained for 150 training stages.
\item Categorical crossentropy as loss function.
\item ADAM optimizer.
\end{itemize}

The hyperparameters for the algorithmic weighting function used were:
\begin{equation}\label{wf}
f(x_i)=\begin{cases}
0.01 &\text{if } BDM(x_i) \in Q(75, BDM(C(x_i))) \\
0.5 &\text{if }  BDM(x_i) \in Q(50, BDM(C(x_i)))\\
2 &\text{if } BDM(x_i) \in Q(0, BDM(C(x_i))),
\end{cases}
\end{equation}
which means that if the BDM value for the $i$-th sample was in the 75th quantile of the algorithmic complexity within its class, then it was assigned a weight of 0.01; the assigned weight was 0.5 if it was in the 50th quantile, and 2 if it was among the lower half in terms of algorithmic complexity within its class. The value for these hyperparameters was found by \textit{random search}. That is, we tried various candidates for the function on the validation set and we are reporting the one that worked best. Although not resorted to for this particular experiment, more efficient hyperparameter optimization methods such as grid search can be used.

Following the theoretical properties of algorithmic regularization, by introducing algorithmic probability weighting we expected to steer the fitting of the target parameters away from random noise and towards the regularities found in the training set. Furthermore, the convergence toward the minimum of the loss function is expected to be significantly faster, in another instance of algorithmic probability \textit{speed-up} (\cite{hernandez2018algorithmically}). We expected the positive effects of the algorithmic probability weighting to increase with the variance of the model to which it was applied. This expectation confirms the hypothesis of the next numerical experiment.

\begin{figure}
\subfloat{%
       \includegraphics[width=7cm]{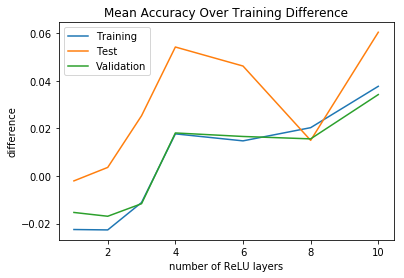}
     }
\hfill 
\subfloat{%
       \includegraphics[width=7cm]{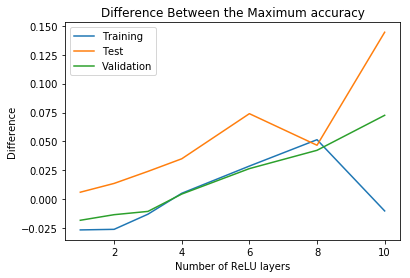}
     }

\subfloat{%
       \includegraphics[width=7cm]{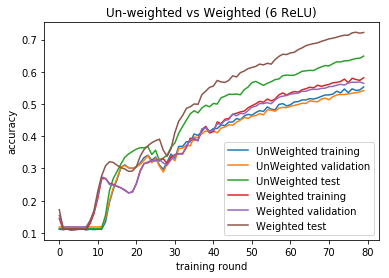}
     }
\hfill 
\subfloat{%
       \includegraphics[width=7cm]{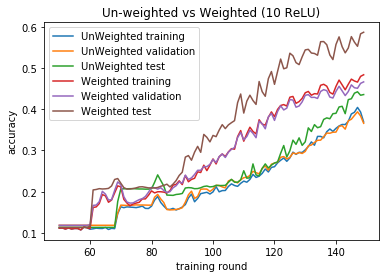}
     }
     

\caption{The first two (upper) plots show the difference between the mean and maximum accuracy obtained through the training of each of the models. The last two (lower) plots show the evolution of accuracy through training for the data sets. The data sets used are (\texttt{training, test} and \texttt{validation}), with data from the MNIST dataset.  The (\texttt{training} and \texttt{validation}) data sets were salted with \%40 of the data randomly corrupted while the \texttt{test} set was not.  From the first two plots we can see that the accuracy of the models trained with algorithmic sample weights is consistently higher than the models trained without them, and this effect increases with the variance of the models. The drops observed after 4 ReLU layers are because, until depth 10, the number of training rounds was constant, with more training rounds therefore needed to achieve a minimum in the cost function. When directly comparing the training history of the models of depth 6 and 10 we can see that the stated effect is consistent. Furthermore, at 10 rectilinear units, we can see significant overfitting, while for the unweighted model, using the algorithmic weights still leaves room for improvement.}\label{accuracy}
\end{figure}

\begin{figure}[ht!]
    \centering
    \includegraphics[width=11cm]{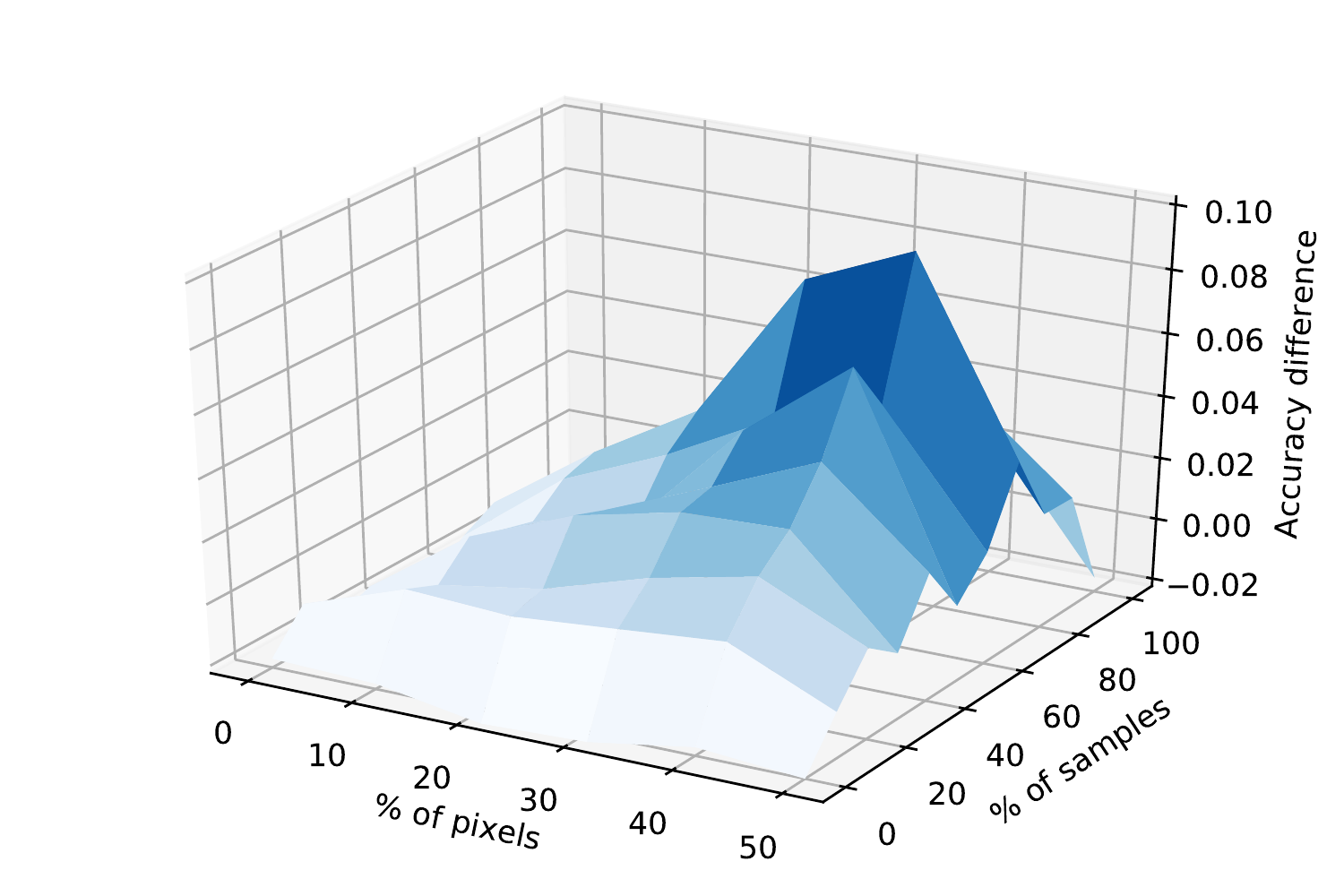}
    \caption{The difference in accuracy with respect to the percentage of corrupted pixels and samples in the data set for the weighing function \ref{wf} for a neural network of depth 4 (four rectilinear units). A positive value indicates that the network trained on the weighted samples reached greater accuracy. The maximum difference was reached for 70\% of the samples with 40\% of pixels corrupted. From the plot we can see that the networks trained over the weighed samples steadily gained in accuracy until the maximum point was reached. The values shown are the average differences over five networks trained over the same data.}
    \label{wg}
\end{figure}

\begin{figure}[ht!]
\subfloat{%
       \includegraphics[width=7.4cm]{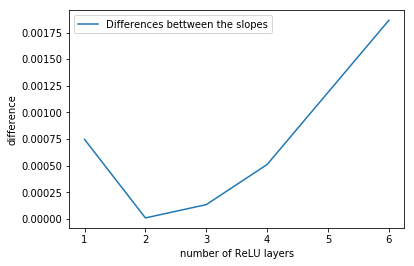}
     }
\hfill 
\subfloat{%
       \includegraphics[width=7cm]{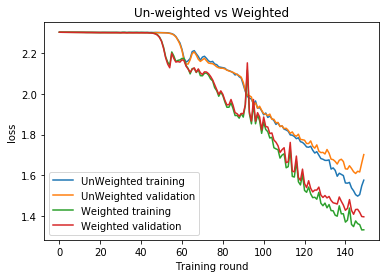}
     }

\caption{On the left are shown the differences between the slopes of the linear approximation to the evolution of the loss function for the first six weighted and unweighted models. The linear approximation was computed using linear regression over the first 20 training rounds. On the right we have the loss function of the models with 10 ReLU units. From both plots we can see that training toward the minimum of the loss function is consistently faster on the models with the algorithmic complexity sample weights, and that this difference increases with the variance of the model.}\label{loss}
\end{figure}

The differences in the accuracy of the models observed through the experiments as a function of variance (number of ReLU layers) are summarized in Figure~\ref{accuracy}. The upper plots show the difference between the mean accuracy and the maximum accuracy obtained through the optimization of the network parameters for networks of varying depth. A positive value indicates that the networks trained with the algorithmic weights showed a higher accuracy than the unweighted ones. The difference in the \textit{steepness} of the loss function between the models is shown in the left plot of Figure~\ref{loss}, which is also presented as a function of the number of ReLU layers. A positive value indicates that a linear approximation to the loss function had a steeper gradient for the weighted models when compared to the unweighted ones. In Figure~\ref{wg}, we can see the evolution of this difference with respect to the percentages of corrupted samples and the corrupted pixels within these samples. 

As the data show (Figure~\ref{accuracy}), the networks trained with the algorithmic weights are more accurate at classifying all three sets: the salted training set, the (unsalted) test set and the (salted) validation set. This is shown when the difference of the mean accuracy (over all the training epochs) and the maximum accuracy attained by each of the networks is positive. Also, as predicted, this difference increases with the variance of the networks: at higher variance, the difference between the accuracy of the data sets increases. Moreover, as shown in Figure~ \ref{loss}, the weighted models reach the minimum of the loss function in a lower number of iterations, exemplified when the linear approximation to the evolution of the cost is steeper for the weighted models. This difference also increases the variance of the model.

\section{Conclusions}

Here we have presented a mathematical foundation within which to solve supervised machine learning tasks using algorithmic information and algorithmic probability theories. We think this is the first time that a symbolic inference engine is integrated to more traditional machine learning approaches constituting not only a path towards putting both symbolic computation and statistical machine learning together but allowing a state-to-state and cause-and-effect correspondence between model and data and therefore a powerful interpretable white-box approach to machine learning. This framework is applicable to any supervised learning task, does not require differentiability, and is \textit{naturally biased} against complex models, hence inherently designed against over-fitting, robust in the face of adversarial attacks and more tolerant to noise in continuous identification problems.

We have shown specific examples of its application to different problems. These problems included the estimation of the parameters of an ODE system, the classification of the evolution of elementary cellular automata according to their underlying generative rules; the classification of binary matrices with respect to 10 initial conditions that evolved according to a random elementary cellular automaton; and the classification of the evolution of a Boolean NK network with respect to 10 associated binary rules or ten different network topologies, and the classification of the evolution of a randomly chosen network according to its connectivity (the parameter $k$). These tasks were chosen to highlight different approaches that can be taken to applying our model. We also assert that for these tasks it is generally hard for non-specialized classifiers to get accurate results with the amount of data given.

While simple, the ODE parameter estimation example illustrates the range of applications even in the context of a simple set of equations where the unknown parameters are those explored above in the context of a neural network~\cite{dua2011simultaneous}, $[\theta_1 \; \theta_2]=[5 \; 1]$. These parameters correspond to a low algorithmic complexity model. Given the way that algorithmic parameter optimization works, the optimization time, as measured by the number of iterations, will converge faster if the optimal parameters have low algorithmic complexity, and therefore are more plausible in the algorithmic sense. These low complexity assumptions are compatible with a principle of parsimony such as Ockham's razor, empirically found to be able to explain data simplicity bias~\cite{algo,kamal,zenilbadillo}, and suggesting that the best explanation is also the simplest, but also that what is modelled is not algorithmically random~\cite{zenilchaitin}. The advantage of our approach is that it offers a means to reveal a set of candidate generative models.

From the results obtained from the first classification task (\ref{fECA}), we can conclude that our vanilla algorithmic classification scheme performed significantly better than the non-specialized vanilla neural network tested. For the second task (Section \ref{fI}), our algorithmic classifier achieved an accuracy of 95.5\%, which was considerably higher than the 60.11\% achieved by the best performing neural network tested.

For finding the underlying topology and the Boolean functions associated with each node, the naive neural network achieved a performance of 92.50\%, compared to 91.35\% for our algorithmic classifier. However, when classifying with respect to the topology, our algorithmic classifier showed a significant difference in performance, with over 39.75\% greater accuracy. There was also a significant difference in performance on the fourth task, with the algorithmic classifier reaching an accuracy of 70\%, compared to the 43\% of the best neural network tested.

We also discussed some of the limitations and challenges of our approach, but also how to combine and complement other more traditional statistical approaches in machine learning. Chief among them is the current lack of a comprehensive Turing machine based conditional CTM database required for the strong version of conditional BDM. We expect to address this limitation in the future.

It is important to emphasize that we are not stating that there is no neural network that is able to obtain similar, or even better, results than our algorithms. Neither do we affirm that algorithmic probability classification in its current form is better on any metric than the existing extensive methods developed for deep learning classification. However, we have introduced a completely different view, with a new set of strengths and weaknesses, that with further development could represent a better grounded alternative suited to a subset of tasks beyond statistical classification, where finding generative mechanisms or first principles are the main goals, with all its attendant difficulties and challenges.

\bibliography{ref}


\newpage

\section{Appendix}\label{app}

\subsection{Joint and Mutual BDM} \label{jointAndMutual}
In classical information theory (\cite{entropy, cover2012elements}) we can think of \textit{mutual entropy} as the information contained over two or more events occurring concurrently, and of \textit{joint entropy} over the two communication channels or events as the average uncertainty contained over all possible combinations of events. For algorithmic information theory, the first concept can be understood as the ``\textit{amount of information within an object that is explained by another}'' and the second concept can be interpreted as the ``\textit{amount of information contained within two or more objects}''.

In contrast to classical information theory, we started by defining conditional BDM. Therefore we think that the best way to define joint BDM is from the \textit{chain rule}.

\begin{defn}\label{jointBDM}
The \emph{joint BDM of $X$ and $Y$ with respect to $\{\alpha_i\}$} is defined as

\begin{equation*}
JointBDM(X,Y) = BDM(Y|X) + BDM(X).
\end{equation*}
\end{defn}

Following the same path, we could define \textit{mutual BDM} thus:

\begin{defn}\label{mutualBDM}
The \emph{mutual BDM of $X$ and $Y$ with respect to $\{\alpha_i\}$} is defined as
\begin{equation*}
MutualBDM(X,Y) = BDM(X) - BDM(X|Y).
\end{equation*}
\end{defn}

\subsubsection{The Relationship Between Conditional, Joint and Mutual Information}\label{properties}

The results shown in this section are evidence that our Def.~for conditional BDM is \textit{well behaved}, as it is analogous to important properties for conditional, joint and mutual entropy.

\begin{prop}
If $X=Y$ then $BDM(X|Y)=0$.
\begin{proof}
is a direct consequence of the Def.~\ref{conditionalBDM}.
\end{proof}
\end{prop}

It is important to note that $BDM(X|Y)=0$ does not imply that $X=Y$. However, it does imply that $Adj(X)=Adj(Y)$. This is a consequence of the fact that BDM does not measure the information encoded in the position of the subtensors.

\begin{prop}
 $BDM(X) \geq BDM(X|Y)$.
\begin{proof}
As we consider subsets of $Adj(X)$, it is a direct consequence of the Def.~\ref{conditionalBDM}.
\end{proof}
\end{prop}

\begin{prop}
If $X$ and $Y$ are independent with respect to the partition $\{\alpha_i\}$, this is equivalent to $Adj(X) \cap Adj(Y)=\emptyset$, then $BDM(X|Y)=BDM(X)$.
\begin{proof}
It is a direct consequence of the Def.~\ref{conditionalBDM}, given that we have it that $Adj(X)-Adj(Y)=Adj(X).$
\end{proof}
\end{prop}

\begin{prop}\label{prop1}
$MutualBDM(X,Y) = MutualBDM(Y,X)$.

\begin{proof}
First, consider the equation
\begin{align*}
MutualBDM(X,Y) = & BDM(X) - BDM(X|Y) \\
= & \sum_{(r_i,n_i) \in  Adj(X)} CTM(r_i)+\log (n_i)\\
& - \sum_{(r_i,n_i) \in  Adj(X)-Adj(Y)}( CTM(r_i)+\log (n_i) ) \\
&- \sum_{Adj(X) \cap Adj(Y)} f(n^x_k,n^y_k).\\
\end{align*}

\noindent{}While on the other hand we have it that 
\begin{align*}
MutualBDM(Y,X) = & BDM(Y) - BDM(Y|X) \\
= & \sum_{(r_j,n_j) \in  Adj(Y)} CTM(r_j)+\log (n_j)\\
& - \sum_{(r_j,n_j) \in  Adj(Y)-Adj(X)}( CTM(r_j)+\log (n_j) ) \\
& - \sum_{Adj(Y) \cap Adj(X)} f(n^y_k,n^x_k).\\
\end{align*}
Notice that in both equations we have the sum over all the pairs that are in both sets, $Adj(X)$ and $Adj(Y)$, with the difference being in the terms corresponding to the \textit{multiplicity}. Now we have to consider two cases. If $n^x_i=n^y_i$ we have the equality. Otherwise, in the first equation we have terms of the form $\log (n^x_j)-f(n^x_k,n^y_k)$, which, by Def.~of $f$, is 0; analogously for the second equation. Therefore, we have the equality. 
\end{proof}

\end{prop}

\begin{prop}\label{prop2}
$MutualBDM(X,Y) = BDM(X) + BDM(Y) - JointBDM(X,Y)$.

\begin{proof}

\begin{align*}
MutualBDM(X,Y) & = MutualBDM(Y,X)\\
& = BDM(Y)  - BDM(Y|X)\\
& = BDM(Y) + BDM(X)  - (BDM(Y|X) + BDM(X))\\
& = BDM(X) + BDM(Y) - JointBDM(X,Y)
\end{align*}
\end{proof}

\end{prop}

\subsection{Coarseness and Relationship With Entropy}\label{coarseness}

As mentioned in the previous section, the goal behind the Def.~of coarse conditional BDM, $BDM(X|Y)$, is to measure the amount of information contained in $X$ not present in $Y$. Ideally, this is measured by the conditional algorithmic information $K(X|Y)$. The Def.~\ref{conditionalBDM} includes the adjective  \textit{coarse} given that, as we will show in this section, its behaviour is closer to Shannon's entropy $H$ than the algorithmic information measure $K$, relying heavily on the entropy-like behaviour of BDM.

The conditional algorithmic information content function $K$ is an incomputable function. Therefore it represents a theoretical ideal that cannot be reached in practice. By construction, \textit{coarse} conditional BDM is an approximation to this measure. However it differs in not taking into account two information sources: the information content shared between base blocks and the position of each block.

As an example of the first limitation, consider the string $101010\dots10$ and its \textit{negation} $010101\dots01$. Intuitively, we know that both strings are algorithmically close, but for a partition strategy that divides the string into substrings of size 2 with no overlapping, the $Adj$ sets $\{(\{10\}, n)\}$ and  $\{(\{01\}, n)\}$ are disjoint. Therefore conditional BDM assigns the maximum BDM value to the shared information content. Within this limitation, we argue that conditional BDM represents a better approximation to $K$ in comparison to entropy, mainly because BDM uses the CTM approximation value for each block, rather than just its distribution, and the information content of its multiplicity, thus representing a more accurate approximation to the overall algorithmic information content of the non-shared base blocks.\\

The second limitation can become a significant factor when the size of the base blocks is \textit{small} when compared to that of the objects analysed, given that the positional information can become the dominant factor of the information content within an object. This is an issue shared with entropy that conditional BDM inherits from the numerical challenges of CTM in BDM. However, conditional BDM has the added benefit that it is defined for finite tensors generated from different distributions by assuming the so-called \textit{universal distribution} (\cite{solomonoff2003kolmogorov}) (known to dominate any other approach) as the underlying distribution between the two `\textit{events}'.

\subsubsection{Empirical Comparison with Entropy}\label{HvsBDMSec}

Owing to the origins of the BDM function, the asymptotic relationship between coarse conditional BDM and conditional entropy follows from the relationship between BDM and entropy (\cite{bdm}). In this section we will focus on empirical evidence for this relationship, along with exploring the impact of the partition strategy for unidimensional objects. Further theoretical properties that establish the \textit{well-behavedness} of conditional BDM are set forth in the Appendix in Section~\ref{jointAndMutual}.

\begin{figure}
\centering
\includegraphics[width=0.5\textwidth]{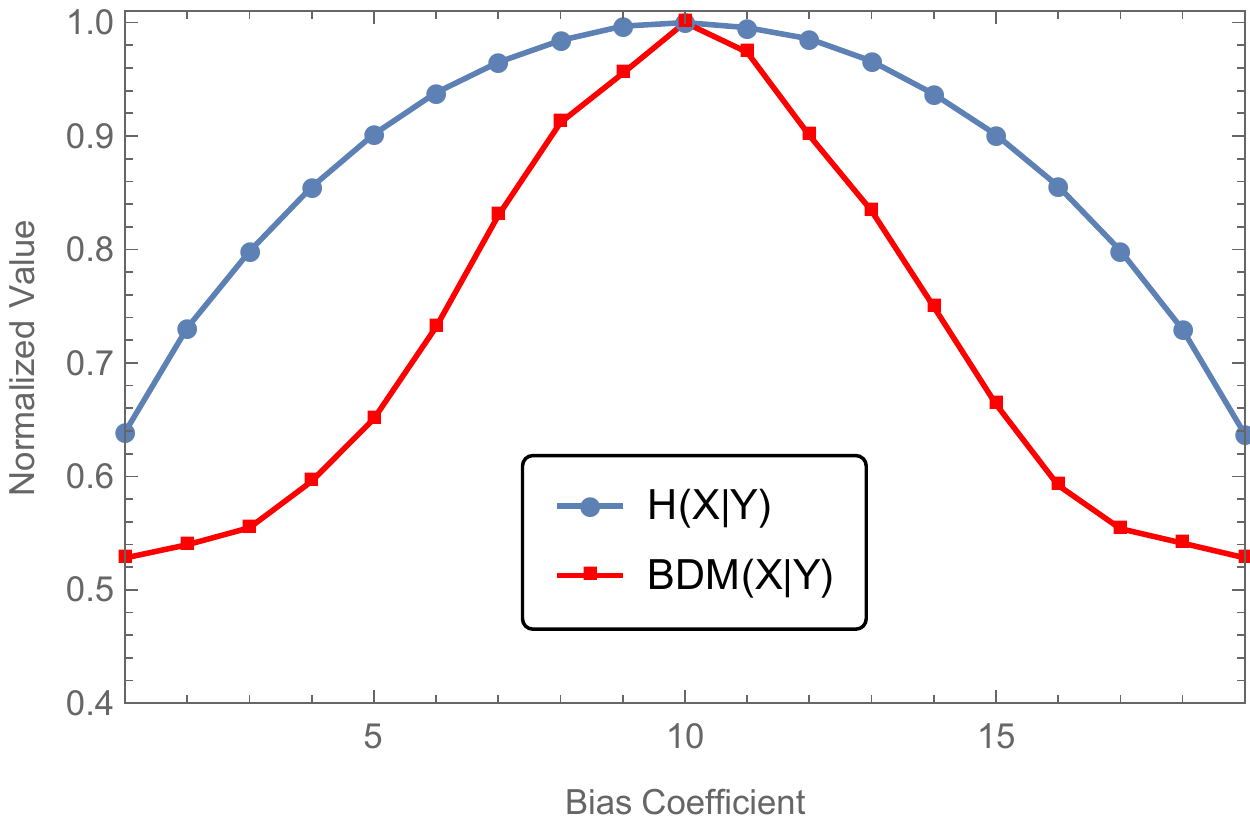}
\caption{\label{HvBDM} Each point represents the normalized average of the conditional BDM ($BDM(X|Y)$) and conditional entropy ($H(X|Y)$), respectively, corresponding to 5000 pairs of strings randomly chosen from a distribution where the expected number of 1s is the value shown on the \emph{x axis} divided between the conditional BDM or conditional entropy of the first element of the pair and an unrelated randomly chosen binary string. All strings are of length 20. The partition strategy used for BDM is that of sets of size 1. From this plot we can see that conditional BDM manages to capture the statistical relationship of finite strings generated from the same distribution.}
\end{figure}

For this numerical experiment we generated a sample of 19,000 random binary strings of length 20 that are pairwise related, coming from one of 19 \textit{biased} distributions where the expected number of 1s varies from 1 to 19. For each pair we computed the conditional BDM with partitions of size 1 and divided it by the conditional BDM of the first string with respect to a random string coming from an uniform distribution. To both, the divisor and the dividend, we added 1 to avoid divisions by zero. We repeated the experiment for conditional entropy. Both results where normalized by dividing the quotients obtained by the maximum value obtained for each distribution. In the plot \ref{HvBDM} we show the average obtained for each biased distribution.\\

From the plot \ref{HvBDM} we can see that as the underlying distribution associated with the strings is increasingly biased, the expected shared information content of two related strings is higher (conditional BDM is lower) when compared to the conditional BDM of two unrelated strings. This behaviour is congruent with what we expect and observe for conditional entropy. That the area under the normalized cube is smaller is expected, given that BDM is a finer-graded information content measure than entropy and is not perfectly symmetric, as BDM and CTM are computational approximations to an uncomputable function and are also inherently more sensitive to the fundamental limits of computable random number generators.

\subsection{The Impact of the Partition Strategy}

As shown in previous results (\cite{bdm}), BDM better approximates the universal measure $K(X)$ as the number of elements resulting from applying the partition strategy $\{\alpha_i\}$ to $X$. However, this is not the case for conditional BDM. Instead $BDM(X|Y)$ is a good approximation to $K(X|Y)$ when the $Adj(X)$ and $Adj(Y)$ share a high number of base tensors, and the probability of this occurring is lower in inverse proportion to the number of elements of the partition. For this reason we must point out that conditional BDM is dependent on the chosen partition strategy $\{\alpha_i\}$.\\

As a simple example, consider the binary string $X=11110000$ and its inverse $Y=00001111$. Since we have the CTM approximation for strings of size 8, the best BDM value for each string is found when $Adj(X)=\{(11110000,1)\}$ and $Adj(Y)=\{(00001111,1)\}$. However, given that the elements of the partitions are different, we have it that $BDM(11110000|00001111)=BDM(11110000)=25.1899,$, even when intuitively we know that, algorithmic information-wise, they should be very close. However, conditional BDM is able to capture this with partitions of size 1 to 4 with no overlapping, assigning a value of 0 to $BDM(X|Y)$.\\

We conjecture that there is no general strategy for finding a \textit{best partition strategy}. This is an issue shared with conditional block entropy, and just like the original BDM definition. At its worst, conditional BDM will behave like conditional entropy when comparable, while maintaining best cases close to the ideal of conditional algorithmic complexity. Thus the partition strategy can be considered an \textit{hyperparameter} that can be empirically optimized from the available data.

\begin{figure}
\centering
\includegraphics[width=0.5\textwidth]{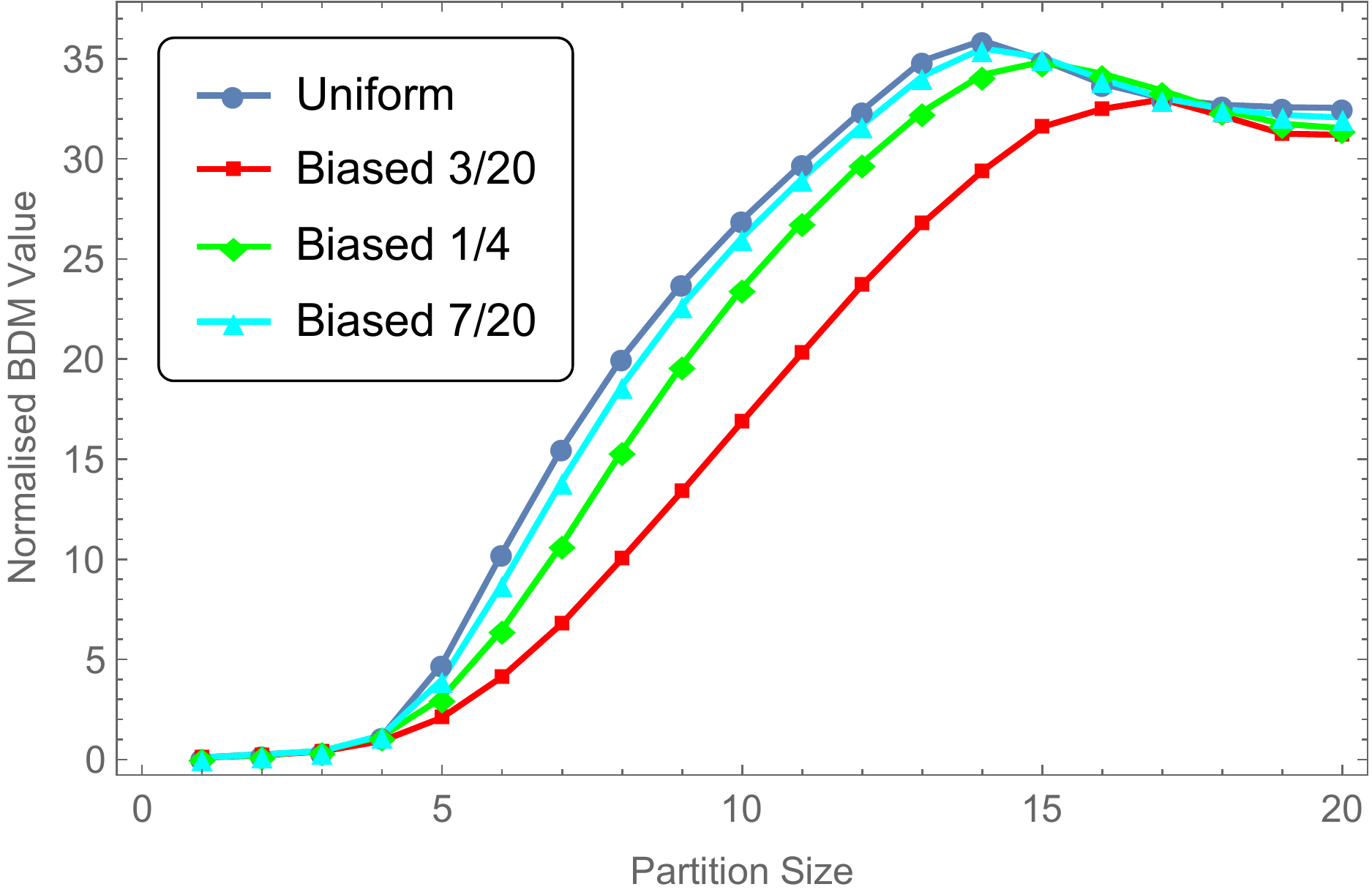}
\caption{\label{part}Each point represents the average of the conditional BDM from 30,000 pairs of binary strings of size 20 randomly generated from four different distributions: \textit{uniform} (ten 1s expected), \textit{biased 3/20} (three 1s expected), \textit{biased 1/4} (five 1s expected) and \textit{biased 7/20} (seven 1s expected). The \textit{x axis} indicates the partition size used to compute the respective conditional BDM value, which was normalized by dividing it by the partition size. 
}
\end{figure}

We performed a numerical experiment to observe this behaviour by generating 2\,400\,000 random binary strings of size 20 with groups of 600,000 strings belonging to one of four different distributions: \textit{uniform} (ten 1s expected), \textit{biased 3/20} (three 1s expected), \textit{biased 1/4} (five 1s expected) and \textit{biased 7/20} (seven 1s expected). Then, we formed pairs of strings belonging to the same distribution and computed the conditional BDM using different partition sizes from 1 to 20, for a total of 30,000 pairs per data point, normalizing the result by dividing it by the partition size to avoid this factor being the dominant one. In the plot \ref{part} we show the average obtained for each data point.\\

In figure~\ref{part} we can observe two main behaviours. The first is that as the partition size increases so does the conditional BDM value. This is because bigger partitions take into account more information from the position of each bit, and we do not expect randomly generated strings to share positional information. The drop observed after partitions of size 12 is the result of CTM values being available up to strings of size 12, the point where the program starts to rely on BDM for the computation. Additionally the partition strategy ignores smaller partitions than the ones stated, thereby reducing the overall amount of information taken into account.
The second is that not only is conditional BDM able to capture the discrepancies expected from the different distributions for partition sizes where there is no loss of statistical information (this being from size 1 to 10), but seems to improve on its ability to do so with larger partition sizes up to 10, therefore improving upon the results presented in Section~\ref{HvsBDMSec}.

It is important to note that an important reduction in accuracy for partitions of sizes larger than 10 was expected, given that the partition strategy used discarded substrings of smaller sizes than the ones stated. For instance, the partition of size 3 of the string 10111 is just $\{101\}$, thus losing information. Furthermore, for big partition sizes with respect to the string length, the statistical similarity vanishes, given that now each substring is considered a \textit{different symbol of an alphabet}. Therefore, the abrupt change of behaviour observed beyond partitions of size 15 is expected and is the product of causation.

\subsection{Experiments and models}

\begin{equation}\label{model}
M=\left\{
\begin{array}{rcrc}
       \text{704} & \rightarrow \{0, 0, 1, 0, 1, 1, 0, 0, 0, 0, 0, 0\},  & 3572&\rightarrow \{1, 1, 0, 1, 1, 1, 1, 1, 0, 1, 0, 0\},\\
       \text{3067} & \rightarrow \{1, 0, 1, 1, 1, 1, 1, 1, 1, 0, 1, 1\},  & 3184&\rightarrow \{1, 1, 0, 0, 0, 1, 1, 1, 0, 0, 0, 0\},\\
       \text{1939} & \rightarrow \{0, 1, 1, 1, 1, 0, 0, 1, 0, 0, 1, 1\},  & 2386&\rightarrow \{1, 0, 0, 1, 0, 1, 0, 1, 0, 0, 1, 0\},\\
       \text{2896} & \rightarrow \{1, 0, 1, 1, 0, 1, 0, 1, 0, 0, 0, 0\},  & 205&\rightarrow \{0, 0, 0, 0, 1, 1, 0, 0, 1, 1, 0, 1\},\\
       \text{828} & \rightarrow \{0, 0, 1, 1, 0, 0, 1, 1, 1, 1, 0, 0\},  & 3935&\rightarrow \{1, 1, 1, 1, 0, 1, 0, 1, 1, 1, 1, 1\}\\
\end{array}\right\}
\end{equation}

\subsection{Other result details}\label{detailed}
\begin{table}
  \caption{Accuracy for the first task}
  \label{resTables2}
  \centering
  \begin{tabular}{ccc}
    \toprule
    Classifier    & Accuracy on \texttt{Test Set}  & Accuracy on \texttt{Training Set}  \\
    \midrule
     \multicolumn{1}{c}{Simple Networks}                   \\
    \cmidrule(r){1-3}
    1 & 60.11\% & 98.86\% \\
    2 & 57.30\% & 98.86\\
    5 & 25.84\% & 32.95\%\\
    \cmidrule(r){1-3}
    Fernandes & 18.54\% & 50.56\%\\
    \cmidrule(r){1-3}
    Algorithmic Class. & 95.50\% & 96.02\%\\
    \bottomrule
  \end{tabular}
\end{table}

\begin{table}
  \caption{Accuracy for the second task}
  \label{resTables3}
  \centering
  \begin{tabular}{ccc}
    \toprule
    Classifier    & \texttt{Rules Test Set}  & \texttt{Topology Test Set}  \\
    \midrule
    Logistic Regression & 82.35\% & 20.75\% \\
    NN & 92.50\% & 32.75\% \\
    \cmidrule(r){1-3}
    Algorithmic Class. & 91.35\% & 72.4\%\\
    \bottomrule
  \end{tabular}
\end{table}

\begin{table}
  \caption{Accuracy for the third task}
  \label{resTables4}
  \centering
  \begin{tabular}{ccc}
    \toprule
    Model    &  \texttt{Test Set}  &  \texttt{Training Set}  \\
    \midrule
    \multicolumn{1}{c}{Neural Networks}                   \\
    \cmidrule(r){1-3}
    Naive & 43\% & 100\% \\
    Convolutional & 31.66\% & 97.66\% \\
    \cmidrule(r){1-3}
    Boosted Trees & 35\% & 64.33\%\\
    \midrule
    \multicolumn{1}{c}{Algorithmic Classifiers} \\
    \cmidrule(r){1-3}
    BDM & 70\% & 71\%\\
    Entropy & 37.66\% & 46.33\%\\
    \bottomrule
  \end{tabular}
\end{table}

\end{document}